\documentclass[preprint,12pt]{elsarticle}

\usepackage{amssymb}
\usepackage{longtable}

\usepackage{graphicx}
\usepackage{epstopdf}

\makeatletter
\g@addto@macro\endfrontmatter{\enlargethispage{-2\baselineskip}}
\makeatother

\journal{Elsevier's Neural Networks}

\begin{document}

\begin{frontmatter}



% \title{An Updated and Comprehensive Image Dehazing Review: From Benchmarked Datasets to Remote Sensing and UAV-based Applications}

% \title{Enhancing Vision in the Skies: A Review of Image Dehazing for Remote Sensing and UAV Imagery}
\title{Dehazing Remote Sensing and UAV Imagery: A Review of Deep Learning, Prior-based, and Hybrid Approaches}


\author[label1,label4]{Gao Yu Lee}
%\email{GAOYU001@e.ntu.edu.sg}

\author[label1]{Jinkuan Chen}
%\email{jinkuan.c@gmail.com}

\author[label2]{Tanmoy Dam}
%\email{tanmoydam@yahoo.com}

\author[label3]{Md Meftahul Ferdaus}
%\email{mferdaus@uno.edu}

\author[label1]{Daniel Puiu Poenar}
%\email{EPDPuiu@ntu.edu.sg}

\author[label4]{Vu N.Duong}
%\email{vu.duong@ntu.edu.sg}

\affiliation[label1]{organization={School of Electrical and Electronic Engineering, Nanyang Technological University},%Department and Organization
            addressline={50 Nanyang avenue}, 
            city={Singapore},
            postcode={639798}, 
            country={Singapore}}

\affiliation[label2]{organization={SAAB-NTU Joint Lab, School of Mechanical and Aerospace Engineering, Nanyang Technological University},%Department and Organization
            addressline={50 Nanyang avenue}, 
            city={Singapore},
            postcode={639798}, 
            country={Singapore}}

\affiliation[label3]{organization={Department of Computer Science, The University of New Orleans},%Department and Organization
            addressline={2000 Lakeshore Drive}, 
            city={New Orleans},
            postcode={70148}, 
            state={LA},
            country={United States}}
            
\affiliation[label4]{organization={Air Traffic Management Research Institute, School of Mechanical and Aerospace Engineering, Nanyang Technological University},
            addressline={50 Nanyang avenue}, 
            city={Singapore},
            postcode={639798}, 
            country={Singapore}}

\begin{abstract}
% High-quality images are crucial in remote sensing and UAV applications, but atmospheric haze can severely degrade image quality, making image dehazing an important research area. Since the introduction of deep convolutional neural networks, numerous approaches have been proposed, and even more have emerged with the development of vision transformers and contrastive/few-shot learning. Simultaneously, papers describing dehazing architectures applicable to various Remote Sensing (RS) domains are also being published. Most current reviews on dehazing focus on approaches evaluated on benchmarked haze datasets, with few extensively exploring RS applications until recently. This review is the first, to our knowledge, to provide comprehensive discussions on both existing and very recent dehazing approaches (as of 2024) on benchmarked and RS datasets, including UAV-based imagery. We also highlight open challenges and issues in the current dehazing literature and propose potential solutions and future research directions.
High-quality images are crucial in remote sensing and UAV applications, but atmospheric haze can severely degrade image quality, making image dehazing a critical research area. Since the introduction of deep convolutional neural networks, numerous approaches have been proposed, and even more have emerged with the development of vision transformers and contrastive/few-shot learning. Simultaneously, papers describing dehazing architectures applicable to various Remote Sensing (RS) domains are also being published. This review goes beyond the traditional focus on benchmarked haze datasets, as we also explore the application of dehazing techniques to remote sensing and UAV datasets, providing a comprehensive overview of both deep learning and prior-based approaches in these domains. We identify key challenges, including the lack of large-scale RS datasets and the need for more robust evaluation metrics, and outline potential solutions and future research directions to address them. This review is the first, to our knowledge, to provide comprehensive discussions on both existing and very recent dehazing approaches (as of 2024) on benchmarked and RS datasets, including UAV-based imagery.
\end{abstract}



\begin{keyword}
Computer Vision (CV) \sep Deep Learning (DL)  \sep Image Dehazing \sep Remote Sensing \sep Unmanned Aerial Vehicle (UAV) imagery
\end{keyword}

\end{frontmatter}


\section{Introduction}
\label{sec:sample1}

Hazy conditions, caused by natural phenomena such as rain and snow, as well as man-made disasters like urban and forest fires, can severely degrade image quality in applications such as photography, surveillance, and remote sensing. This degradation leads to reduced contrast and color shifts, ultimately hindering the performance of computer vision (CV) models and resulting in poor object detection, image classification, and image segmentation outcomes.

As a result, the number of research studies dedicated to extracting crisp, high-quality scenes from hazy photos have increased exponentially over the decades. This field of image processing is known as \emph {image dehazing}. Prior to the widespread applications of deep learning in CV and image processing, image dehazing techniques are mostly dependent on a \emph{prior-based} approach, in which a wide variety of presumptions are applied to a given hazy image to extract and compute its dehazing parameter in a statistical manner. Such approaches usually provide good dehazing outputs and performances in one particular hazy scenario and context, but may not work well for other hazy scenarios and contexts. With deep learning, dehazing parameters can usually be computed in an end-to-end manner and have higher performance metrics than prior-based approaches in a wide variety of hazy scenarios. The Atmospheric Scattering Model (ASM) (\cite{mccartney1976optics}, \cite{narasimhan2003contrast}) is often invoked, which assumes airlight and direct attenuation primarily contribute to the hazy image. Such a model is often the most commonly assumed in many dehazing studies, and can be used for both modelling and generating haze. 

Earlier deep learning-based dehazing often invoked Convolutional Neural Networks (CNNs), and it has only been in recent decades that Vision Transformers (ViTs) have been explored and utilized. Owing to the low local inductive bias of ViTs relative to CNNs (because image patches are processed instead of pixel-by-pixel), the former requires a relatively large amount of training data to attain competitive metric performance. For instance, by training a typical ViT on large-scale pre-trained datasets such as JFT-300M \cite{sun2017revisiting}, the ViT can outperform the CNNs in classification tasks. One of the first studies on ViT dehazing, the DehazeFormer \cite{song2023vision}, attained state-of-the-art (SOTA) performance via training on a relatively large hazy dataset (e.g., SOTS-indoor/SOTS-outdoor/SOTS-mix from RESIDE \cite{li2018benchmarking} and RS-HAZE \cite{song2023vision}). Furthermore, ViTs are often computationally demanding and more architecturally complex than CNNs, and hence without further modifications, may not be suitable for deployment on mobile and edge devices such as Unmanned Aerial Vehicles (UAVs) or autonomous cars.

Concurrently, dehazing approaches that utilize contrastive learning and zero- and one-shot learning are also emerging, although such methods are more often utilized in image classification, segmentation, and object detection. These methods do not require a large-scale hazy-clean image pairs for training unlike the CNN and ViT approaches, and also circumvent the problem of the synthesized haze images being less informative and consistent with that of real-life haze \cite{li2023unsupervised}, which can leads to domain shifting. Unlike the majority of CNNs and ViT approaches, contrastive, zero and one-shot learning approaches utilize unsupervised learning, and zero-shot learning does not require ground-truth (clear) images, instead requiring only a given single hazy image \cite{li2020zero}. Such approaches can benefit from better generalization under scenario of varying haze intensities, which could be true for video tasks (e.g., monitoring forest fires, active volcanoes, rain and snowing).

We realize that there are already review papers on dehazing, both recent and old. For instance, Goyal et.al. \cite{goyal2023recent} provided the most updated review on various dehazing approaches on a myriad of benchmark datasets at the time of our writing. However, the dataset discussed did not include hazy images in the context of remote sensing and aerial mobile platforms like Unmanned Aerial Vehicles (UAVs) and autonomous vehicles. For the former, analyzing hazy images is crucial to assess the degree of damage caused by natural disasters like forest fires, as well as identifying hotspots to assist in firefighting efforts. The latter can potentially aid in autonomous operation in small to moderate rain and snow scenarios (where fog could occur concurrently). Agrawal and Jalal \cite{agrawal2022comprehensive} include the numerical results of various dehazing approaches and assessed their performances in terms of whether haze artifacts remained, whether it was applied in a dense fog scenario, and whether over-enhancement ensued, while simultaneously evaluating the speed of inference of each technique. Gui et.al. \cite{gui2023comprehensive} provides a comprehensive survey and taxonomy of deep learning-based dehazing. The latter also covered a review of contrastive and few-shot dehazing works, and discussed some challenges and open issues that remain in the dehazing domain, such as the impact of dehazing on higher-level vision tasks such as image classification and segmentation. However, similar to Agrawal and Jalal and Goyal et.al., these reviews only covered the performance of the techniques on the benchmark dataset, and did not cover those of remote sensing and imagery from mobile platforms. Furthermore, unlike Agrawal and Jalal, the work does not tabulate the quantitative dehazing metrics for the various methods on the benchmarked datasets, which would make it easier to observe the trends of the effectiveness of the proposed approaches, and see which method gave the best dehazing performance as of current. Finally, we note the existence of a remote sensing-based dehazing review by Liu et.al. \cite{liu2021review}; however, to the best of our knowledge, this is the only one of its kind. Table 1 summarizes some of the existing dehazing review papers, the year they were published, and highlight each of their unique features and gaps. 

%detecting Edge Preservation (EP), Color Distortion (CD), Halo Artifacts (HA), Gradient Inverse Artifacts (GIA), Large Haze Gradient (LHG) and Blocking Artifacts (BA)

\begin{table*}
\centering
\caption{Summary of key review papers for dehazing, the year it was published, highlighting its key features and possible gaps.}
\begin{tabular}{p{3cm}p{0.75cm}p{4.5cm}p{4.5cm}}
\hline
\textbf{Review Paper} & \textbf{Year} & \textbf{Key Features} & \textbf{Gaps} \\
\hline
Fang \cite{fang2022review} & 2022 & Short and succinct explanation of the traditional dehazing methods. Mentioned the common evaluation metrics for traditional dehazing methods. & Absence of quantitative comparison between the mentioned-methods.\\
\hline
Guo et.al. \cite{guo2022imagereview} & 2022 & Comprehensive explanation of each mentioned dehazing methods and its mathematical formulations. Extensive quantitative comparison of dehazing methods to various benchmarked datasets, with numerous evaluation metrics. & Lack of quantitative comparison with the newly introduced ViT-based dehazing methods.\\
\hline
Agrawal $\&$ Jalal \cite{agrawal2022comprehensive} & 2022 & Included numerical results of the State-Of-The-Art dehazing approaches on various benchmarked datasets. Also explored recent metrics for dehazing evaluation with their pros and cons. & Does not include discussions on dehazing approaches for remote sensing and mobile platforms.  \\
\hline
\label{tab:Tables_reviews}
\end{tabular}
\end{table*}

% ======================================= SPLIT TABLE ==========================================

\begin{table*}
\centering
\begin{tabular}{p{3cm}p{0.75cm}p{4.5cm}p{4.5cm}}
\hline
\textbf{Review Paper} & \textbf{Year} & \textbf{Key Features} & \textbf{Gaps} \\
\hline
Gui et.al. \cite{gui2023comprehensive} & 2022 & Covers contrastive and few-shot dehazing approaches, as well as GAN-based approaches. Also highlighted current challenges and issues in the dehazing domain.& Does not include numerical results of the discussed approaches on various benchmarked datasets and discussions on remote sensing and mobile platforms. \\
\hline
An et.al. \cite{an2023comprehensive} & 2023 & Explained the needs for single and multiple atmospheric scattering models. Exhaustive reviews on the single/multiple atmospheric scattering models against existing dehazing techniques. Extensive coverage of datasets for both thin and dense haze. & Lack of numerical results and quantitative comparison of discussed atmospheric scattering models against a dataset. Does not include dehazing approaches on remote sensing and mobile platforms. \\
\hline
Goyal et.al \cite{goyal2023recent} & 2023 & Offered one of the most recent comprehensive reviews of deep learning-based dehazing. Included extensive studies on image dehazing related journals and conference articles from the most prominent online search engines. It also provided an exhaustive analysis on a wide range of image dehazing techniques, numerous datasets, and comprehensive coverage of the evaluation metrics. In-depth explanation of the challenges in image dehazing works. & Lack of quantitative comparisons between the covered image dehazing techniques' computational complexity. Does not include dehazing approaches for remote sensing and mobile platforms. \\
\hline
\label{tab:Tables_reviews}
\end{tabular}
\end{table*}

In contrast to other review works, our dehazing review is unique in that we covered both deep learning and prior-based dehazing approaches in not only those evaluated on benchmarked datasets, but also those applied to remote sensing and Unmanned Aerial Vehicles (UAVs). In a similar way to that of Gui et.al., we highlighted some open challenges that remain in the aforementioned domain and provided a broad discussion of how to address it. Furthermore, our review also discusses recently proposed prior-based approaches (as of 2023), which has not been the focus of many review papers. We believe that highlighting recent prior-based methods is also essential, despite the popular notion that deep learning-based methods are a more favorable approach, since such works have also demonstrated competing and promising outputs. The existence of such works also indicates that the research on the prior-based dehazing paradigm has not been completely whittled away. In summary, the contributions of our review are as follows.

\begin{itemize}
    \item To the best of our knowledge, our review is one of the few to cover studies on dehazing priors proposed in more recent times (post 2020) and in more details. As mentioned, this is not only to illustrate that research on prior-based dehazing continues to be active despite popular notions, but also to show that such methods can surpass learning-based approaches in the respective dataset without the need for a large image set for training beforehand. This helps to reduce computational demands and complements dehazing for remote sensing applications.
    \item We have provided the most up-to-date review work on dehazing at the time of our writing, extending and covering works that were published in 2024 for all different types of dehazing approaches applied to different benchmarks and selected applications. 
    \item We also covered works beyond the commonly benchmarked dehazing dataset, to include those utilized in remote sensing and UAVs imagery. Specifically, for the former, we explored dehazing as applied to hyperspectral, the very high resolution (VHR), and synthetic aperture radar (SAR) images. To the best of our knowledge, our work is one of the first to highlight all three key domains when reviewing their dehazing literature.
\end{itemize}

\section{Image dehazing: concepts and principles}

In this section, we provide background information on the principles of image dehazing.

There are multiple ways to view haze appearances in an image: One method is to distinguish between real and synthetic haze. The latter can be generated using a fog machine or mathematical model, for which the Atmospheric Scattering Model (ASM) is the most commonly utilized. The ASM is mathematically defined in Equation 1 as

\begin{equation}
    I(x) = J(x)t(x) + A(1-t(x)),
\end{equation}

where $I(x)$ represents the original hazy image, $J(x)$ denotes the dehazed image, $t(x)$ denotes the transmission coefficient which is defined as $t(x) = \textrm{exp}(-\beta d(x))$, where $d(x)$ is the depth value in the particular pixel values $x$, and $A$ is the ambient airlight term. Equation 1 can be rewritten to recover the dehazed image, as follows:

\begin{equation}
    J(x) = \frac{I(x)-A}{t(x)} + A.
\end{equation}

\begin{figure}[hbt!]
    \centering
    \includegraphics[scale=0.50]{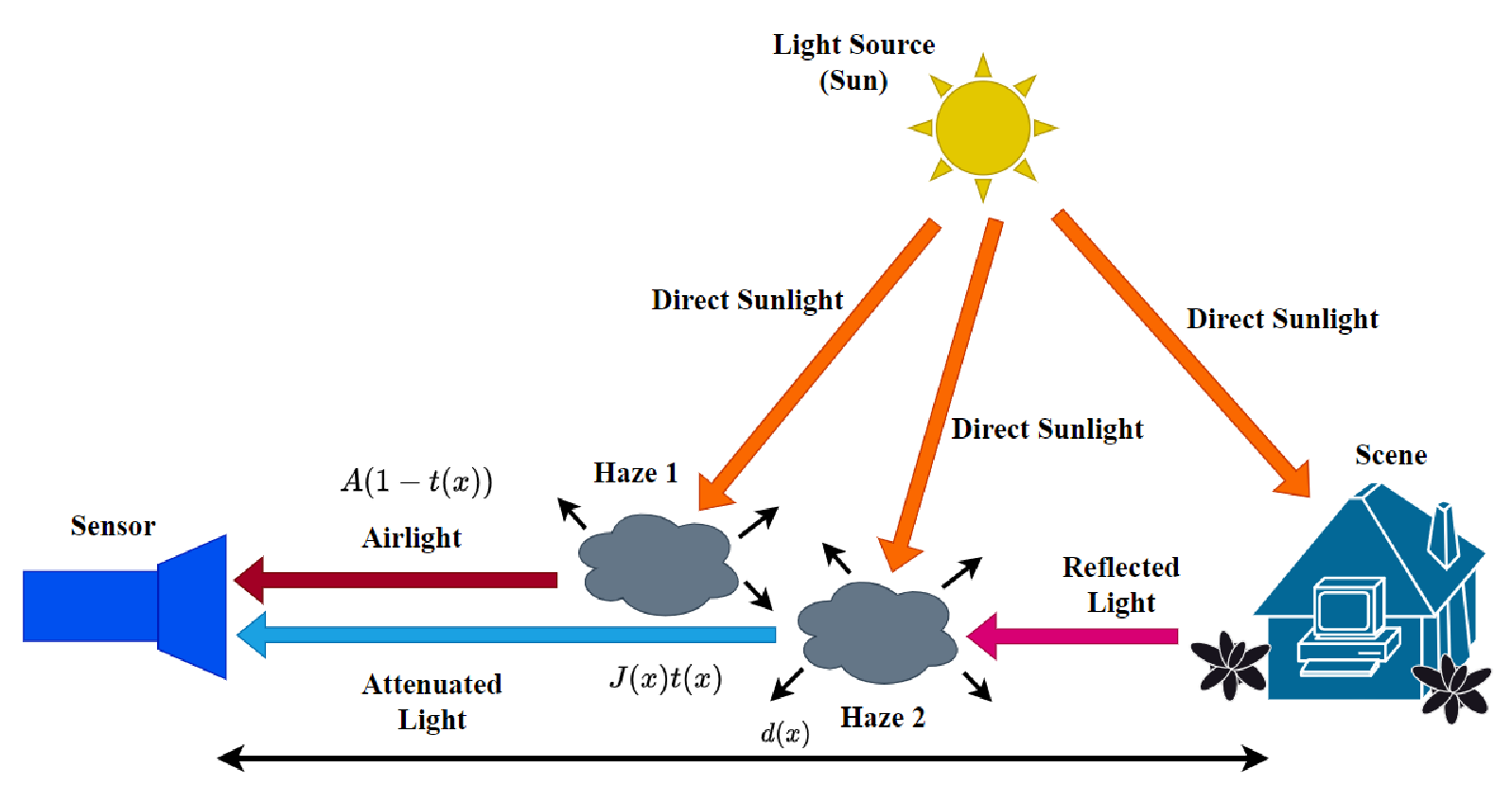}
    \caption{Schematic illustration of the ASM concept. Note that this model is only valid on daytime homogeneous haze. The $J(x)t(x)$ is the direct attenuation term, and is the result of light reflected from a scenery of interest transmitting through a layer of haze. The airlight term $A(1-t(x))$ is the result of light source passing through a layer of haze and illuminating the surrounding. Some levels of light scattering due the haze also contributes to the airlight term. $d(x)$ denotes the distance from the object to the sensor (camera). Blue, maroon, orange, and pink arrows denote attenuated, airlight, direct sunlight, and reflected light respectively.}
    \label{Dehazing_Image}
\end{figure}

Often, in Equation 1, the term $J(x)t(x)$ is also known as the direct attenuation $D(x)$ and the term $A(1-t(x))$ is the airlight $L(x)$ \cite{ancuti2020day}. Figure \ref{Dehazing_Image} depicts a schematic illustration of the principle of ASM, as well as the source of the direct attenuation and airlight term. We can see that the direct attenuation term is the result of light transmitted through a layer of haze, while the airlight term $A(1-t(x))$ is the result of the light source illuminating the surroundings after passing through the haze layer, as well as some levels of light scattering as depicted by the black arrows. 

Image dehazing is often coined as an ill-posed problem, since by looking at equation 1, we can see that there are multiple possible values of $t(x)$ and $A$ that could give the same $I(x)$. The ASM as shown above also assumed that haze is homogeneous and occurs during the daytime. This means that additional modifications must be made for non-homogeneous haze, thick haze, and nighttime haze scenario, for which man-made lights can contribute to the ambient light term $A$. 

Because of their ill-posed nature, prior to the advent of deep learning applications, \emph{prior-based} approaches were usually introduced to estimate the relevant dehazing parameters ($t(x)$ and $A$) by making certain priors, or assumptions, about the hazy image. A limitation of such approaches is that each prior may work well for hazy images in one context, but performs poorly in the others. Hence, learning-based approaches are currently more popular as the dehazing parameters can be computed and inferred by the model directly, sometimes in an end-to-end manner. As described in the introduction, CNNs remain a popular choice for dehazing, although recently, ViTs have been proposed and have shown promising performances relative to those of CNNs.

The majority of dehazing works were tested on benchmarked datasets consisting of synthetic haze images. For example, the RESIDE dataset \cite{li2018benchmarking} remains one of the most popular dehazing benchmark, with 110,500 synthetic hazy images. Another popular benchmark is the D-HAZY dataset \cite{ancuti2016d}, which was adapted from the NYU depth indoor image collection \cite{silberman2012indoor} and comprises more than 1400 real scenery and depth maps for synthetic haze generation. For datasets involving real haze, generation is usually performed using a professional fog machine, and the sample sizes of such datasets is usually much lower than that of synthetic haze. For instance, the O-HAZE dataset \cite{ancuti2018haze} comprises only 55 hazy-clear pairs of real homogeneous haze, NH-HAZE (more specifically NH-HAZE 2020 \cite{ancuti2020nh}) comprises 45 hazy-clear pairs of real non-homogeneous haze, and DENSE-HAZE \cite{ancuti2019dense} consists of 45 hazy-clear pairs of real thick homogeneous haze. Even though the RESIDE dataset contained 4807 real hazy images, it was 23 times smaller than those generated synthetically. Therefore, the metric performances of the majority of dehazing approaches are generally lower for O-HAZE and NH-HAZE, and even more so for DENSE-HAZE, as compared to RESIDE and D-HAZY. 

%Figure 2 depicts sample examples of a hazy image and its corresponding clear image for the various datasets.

%\begin{figure}[hbt!]
    %\centering
    %\includegraphics[scale=0.60]{Dehazing_datasets_image}
    %\caption{Sample examples of a hazy and its corresponding ground-truth clear images from the RESIDE, I-HAZE, O-HAZE, NH-HAZE and DENSE-HAZE datasets, which are some of the most commonly utilized dehazing benchmark.}
    %\label{Dehazing_Image}
%\end{figure}

\section{Commonly utilized dehazing metrics}

Before highlighting some popular approaches to prior-based dehazing and learning-based dehazing (CNNs, ViTs, Contrastive and FSL), we provide a brief summary of the commonly utilized dehazing metrics in a wide variety of literature. 

The most commonly utilized dehazing metrics include the peak-signal-to-noise (PSNR) ratio and the structural similarity index (SSIM) \cite{wang2004image}. The PSNR is defined as the quantitative measure of the extent of restoration between the dehazed image $J(x,y)$ and the target image $K(x,y)$, and is mathematically described as

\begin{equation}
    PSNR = 10 log \left(\frac{max^{2}_{I}}{MSE} \right)
\end{equation}

where $max^{2}_{I}$ is the maximum pixel intensity value possible for both $J(x,y)$ and $K(x,y)$, which in this case is 255, and $MSE$ is the Mean-Squared Error defined as:
\begin{equation}
    MSE = \frac{1}{N} \sum_{(x,y)} (J(x,y) - K(x,y))^{2}.
\end{equation}

Note that $(x,y)$ denotes the pixel coordinates of the image, rather than being associated with depth as in the ASM equation. The PSNR and MSE measures the image quality differences between the dehazed and original clear images in a pixel-by-pixel manner while omitting certain image structures and contexts. The SSIM mitigates this problem and is based on the human visual perception model. This can be described in terms of three key aspects: luminance $l$, contrast $c$, and structure $s$. The SSIM can be written quantitatively as

\begin{equation}
    SSIM = \left[l(J,K)\right]^{\alpha}\left[c(J,K)\right]^{\beta}\left[s(J,K)\right]^{\gamma}
\end{equation}

with $l(J,K)$, $c(J,K)$, and $s(J,K)$ related to relevant means and standard deviations as 

\begin{equation}
l(J,K) = \frac{2\mu_{J}\mu_{K} + C_{1}}{\mu_{J}^{2}\mu_{K}^{2}+ C_{1}},
\end{equation}

\begin{equation}
c(J,K)  = \frac{2\sigma_{J}\sigma_{K} + C_{2}}{\sigma_{J}^{2}\sigma_{K}^{2}+ C_{2}},
\end{equation}
        
\begin{equation}
s(J,K) = \frac{2\sigma_{JK} + C_{3}}{\sigma_{J}\sigma_{K}+ C_{3}},
\end{equation}

where $\mu_{J}$, $\mu_{K}$, $\sigma_{J}$, and $\sigma_{K}$ denote the means and standard deviations of $J(x,y)$ and $K(x,y)$ respectively; $\sigma_{JK}$ is the covariance of $J(x,y)$ and $K(x,y)$, and $C_{1}$, $C_{2}$ and $C_{3}$ are constants set to avoid instability, as highlighted by Bergmann et.al. \cite{bergmann2018improving}.

Some other dehazing metrics utilized include Feature Similarity (FSIM), which was first proposed by Zhang et.al.  \cite{zhang2011fsim}, Natural Image Quality Evaluation (NIQE), which was first introduced by Mittal et.al. \cite{mittal2012making}, and Learned Perceptual Image Patch Similarity (LPIPS), which was first introduced by Zhang et.al. \cite{zhang2018unreasonable}.

FSIM is inspired by the capability of the Human Visual System to understand images based on their low-level features. The FSIM utilizes Phase Congruency (PC), which measures the significance of a local image structure, as well as the Gradient Magnitude (GM), which considers contrast information. The computation of the FSIM was performed in two phases. The first phase involves computation of the local similarity maps between the PCs and GMs, for which the similarity measure ($S_{PC}$) between $PC_{1}$ and $PC_{2}$ is defined as 

\begin{equation}
    S_{PC} = \frac{2PC_{1} \cdot PC_{2} + T_{1}} {PC_{1}^{2} + PC_{2}^{2} + T_{1}},
\end{equation}

while the similarity measure ($S_{GM}$) between $GM_{1}$ and $GM_{2}$ is described as 

\begin{equation}
    S_{GM} = \frac{2GM_{1} \cdot GM_{2} + T_{2}} {GM_{1}^{2} + GM_{2}^{2} + T_{2}}
\end{equation}

where $T_{1}$ is a positive constant to enhance the stability of $S_{PC}$, and the value of $T_{2}$ is dependent on the dynamic range of the GM values. The second phase involves pooling the maps into a single similarity score, and the FSIM score can be obtained by

\begin{equation}
    FSIM = \frac{\sum_{x \in \Omega} S_{L}(x) \cdot PC(x)}{\sum_{x \in \Omega}PC(x)}
\end{equation}

where $S_{L}(x) = \left[S_{PC}(x)\right]^{\alpha}\left[S_{GM}(x)\right]^{\beta}$, and $\Omega$ refers to the fact that the relevant mathematical operation described above is performed over the entire image spatial domain. The FSIM was initially designed for grayscale images or the luminance components of color images, but can be extended to include chromatic information via

\begin{equation}
    FSIM_{c} = \frac{\sum_{x \in \Omega} S_{L}(x) \cdot \left[SC(x)\right]^{\lambda} \cdot PC(x)}{\sum_{x \in \Omega}PC(x)}
\end{equation}

where $SC(x)$ is defined as $S_{I}(x) \cdot S_{Q}(x)$, where $I$ and $Q$ represent the chromatic channels of the respective images.

The NIQE utilizes Natural Scene Statistics (NSS) from a corpus of non-distorted natural scenery and does not require \emph{a priori} exposure to distorted images or any human opinion score data-based training. The NIQE is related to the distance between the NSS feature model and multi-variate Gaussian density as follows:

\begin{equation}
    D(\mu_{1}, \mu_{2}, \Sigma_{1}, \Sigma_{2}) = \sqrt{\left((\nu_{1}- \nu_{2})^{T}\left(\frac{\Sigma_{1} + \Sigma_{2}}{2}\right)^{-1}(\nu_{1} - \nu_{2})\right)}
\end{equation}

where $\nu_{1}$, $\nu_{2}$, $\Sigma_{1}$ and $\Sigma_{2}$ represent the mean vectors and covariance matrices of the natural and distorted Gaussian image distributions, respectively.

The LPIPS computes the similarity between the activation values of  two images using a pre-defined network. It is a perceptual metric, such as SSIM and FSIM, but is designed based on the authors' observation that utilizing deep features for the evaluation metric can outperform all other metrics, which can be extended to a wider variety of deep architectures and supervision types than just the VGG network and supervised learning. A large-scale perceptual similarity dataset was also created to verify their hypothesis, namely the Berkeley-Adobe Perceptual Patch Similarity (BAPPS) dataset.

It is important to note that for PSNR, SSIM and FSIM, the higher their values, the better is the quality of the dehazed images. For NIQE and LPIPS however, the quality of a dehazed image is quantified by how low their values are (with 0 denoting the optimal NIQE and LPIPS value attainable). SSIM, FSIM and LPIPS are bounded between 0 and 1, whereas the MSE, PSNR and NIQE have a lower bound of 0, with no known upper bound. Table 2 summarizes the metrics discussed thus far, covering the year and the authors that contributed, whether it is pixel or perceptual-based, and the range of values it can attain.

\begin{table*} 
\centering
\caption{Summary of key performance metrics for dehazing, including the authors and the year it was published, the range of values it can hold, and whether it is a pixel or perceptual-based metric.}
\begin{tabular}{p{2cm}p{5cm}p{2cm}p{2cm}}
\hline
\textbf{Metrics} & \textbf{Authors and Year} & \textbf{Range of Values} & \textbf{Pixel or Perceptual} \\
\hline
MSE & Known since early times & [0, $\infty$) & Pixel \\
PSNR & Known since early times & [0, $\infty$) & Pixel \\
SSIM & Wang et.al. \cite{wang2004image}, 2004 & [0,1] & Perceptual \\
FSIM & Zhang et.al. \cite{zhang2011fsim}, 2011 & [0,1] & Perceptual \\
NIQE & Mittal et.al. \cite{mittal2012making}, 2012 & [0,$\infty$) & Perceptual \\
LPIPS & Zhang et.al. \cite{zhang2018unreasonable}, 2018 & [0,1] & Perceptual \\
\hline
\end{tabular}
\label{tab:table2}
\end{table*}

\section{Dehazing approaches}

\subsection{Prior-based image dehazing techniques}

One of the most popular approach in this category is the dark channel prior (DCP) proposed by He et.al. \cite{he2010single}. DCP assumes that in a hazy-free image, pixels that do not cover the sky have a minimum of one RGB color channel that has a lower pixel intensity. He et.al. coined these as dark pixels and argued that in hazy images, ambient light $A$ contributes mostly to these dark pixels and can be utilized for direct transmission map estimation to recover the dehazed images. The dark pixels are mathematically defined as 

\begin{equation}
    J^{dark}(x) = \min_{y \in \Omega_{r}(x)} \left(\min_{c \in (r,g,b)} J^{c}(y) \right),
\end{equation}

where $\Omega_{r}(x)$ denotes the small patch centralized at $x$ and with a patch size of $R \times R$. A larger $R$ may cause halo artifacts in the restored images, and the original work approached a trial-and-error paradigm to select the optimal $R$ value of 15.

Fattal \cite{fattal2008single} introduced the Surface Shading Prior (SSP), in which the surface shading function of a scenery in an image is not statically correlated with the scene transmission over a local set of pixels in haze images for dehazing. However, the prior does not incorporate the ASM and hence may introduced color artifacts and contrasts. 

Zhu et.al. \cite{zhu2015fast} presented the Colour Attenuation Prior (CAP), which estimates the transmission depth by assuming that a pixel's brightness and saturation in an unclear environment can be described linearly with regard to depth. This relationship can be mathematically described as follows,

\begin{equation}
    d(x) = \theta_{0}(x) + \theta_{1}(x) \nu (x) + \theta_{2}(x) s(x) + \epsilon(x),
\end{equation}

where $\nu(x)$ is the brightness of the hazy image at a particular pixel, $s(x)$ is the saturation value, $\epsilon(x)$ is a random variable denoting the model's random error, and $\theta_{0}, \theta_{1}$ and $\theta_{2}$ represents the linear coefficients to be estimated. Nevertheless, CAP may produce erroneous estimations in these areas as it is insensitive to landscapes or items with intrinsic white color intensities. 

The Boundary Constraint Context Regularization (BCCR) was introduced by Meng et.al. \cite{meng2013efficient} which improves upon the DCP by estimating the initial values of the transmission maps using inherent boundary constraints. The radiance intensity of a haze-free image $J^{c}$ in a small patch $\Omega$ is constrained in the range $[C_{0}, C_{1}]$, and the initial transmission map is computed as:

\begin{equation}
    t_{b}(x) = \min\left(\max_{(r,g,b)}\left(\frac{A^{c} - I^{c}}{A^{c}-C^{c}_{0}}, \frac{A^{c} - I^{c}}{A^{c}-C^{c}_{1}}\right),1 \right),
\end{equation}

where $C^{c}_{0}$ and $C^{c}_{1}$ denote the RGB channel of $C_{0}$ and $C_{1}$ respectively. The transmission values are assumed to be constant in the small patches ($\omega_{x}$, $\omega_{y}$), and the final transmission map can be computed as 

\begin{equation}
    \bar{t} = \min_{y \in \omega_{x}} \max_{z \in \omega{y}} t_{b}(z).
\end{equation}

The Optimized Contrast Enhancement (OCE) by Kim et.al. \cite{kim2013optimized} utilized the prior that the contrast of a haze-free image is much higher than that of a hazy image. Therefore, the transmission map is estimated via maximizing each patch's contrast. The optimal transmission value in one block is estimated via the minimization of the sum of the two loss function terms: the contrast loss $\ell_{contrast}$ and the information loss $\ell_{info}$, as follows

\begin{equation}
    \min_{t} \left(\ell_{contrast} + \lambda_{L}\ell_{info} \right),
\end{equation}

where $\lambda_{L}$ is the weight value controlling the relative importance between $\ell_{contrast}$ and $\ell_{info}$, which are define mathematically as

\begin{equation}
    \ell_{contrast} = - \sum_{c\in(r,g,b)} \sum_{p\in \Omega} \left(\frac{J^{c}(p) - \bar{J}^{c}}{N_{\Omega}} \right)= - \sum_{c\in(r,g,b)} \sum_{p\in \Omega} \left(\frac{I^{c}(p) - \bar{I}^{c}}{t^{2}N_{\Omega}} \right),
\end{equation}

\begin{equation}
    \ell_{info} =  \sum_{c\in(r,g,b)} \sum_{p\in B} \left(\min(0,J^{c}(p))\right)^{2} + \sum_{c\in(r,g,b)} \sum_{p\in B} \left(\max(0,J^{c}(p)-255)\right)^{2},
\end{equation}

where $N_{\Omega}$ shown in Equation 8 denotes the number of pixels in patch $\Omega$, and the respective quantities with a bar on top denote its average value in that patch. In the original OCE study, the patch size selected was $32 \times 32$. 

In the non-Local image dehazing (NLD) by Berman et.al. \cite{berman2016non}, the assumptions made were that the color of a haze-free image were consisted of few individual colors coined as a haze line. It deviates from the aforementioned patch-based methods, with the modified hazy image $I_{A}(x)$ expressed as

\begin{equation}
    I_{A}(x) = I(x) - A,
\end{equation}

noting that the original RGB coordinates are translated such that the airlight term is now at the coordinate origin. The color of the pixels is represented in a spherical coordinate $(r(x), \theta(x), \phi(x))$ system around the airlight, and the transmission based on the radii in the haze line is expressed as

\begin{equation}
    t(x) = \frac{r(x)}{r_{max}},
\end{equation}

where $r(x) = t(x)||J - A||$, $0 \leq t(x) \leq 1$,  and $r_{max} = ||J - A||$, corresponding to the largest radii line of $t = 1$.

All the aforementioned methods were proposed and published pre-2020. As described in the previous section, prior-based approaches have been researched and proposed currently. For instance, Li et.al. \cite{li2023single} in 2023 introduces an improved bright channel and dark channel prior, which first segmented hazy images into sky and non-sky regions via Otsu's method with particle swarm optimization, followed by utilizing the bright channel prior (BCP) to estimate the parameters of the bright pixels as

\begin{equation}
    J^{bright}(x) = \max_{y \in \Omega_{r}(x)} \left(\max_{c \in (r,g,b)} J^{c}(y) \right),
\end{equation}

noting that the BCP has the same mathematical form as DCP but with $min$ replaced with $max$. The above equation is used to estimate the ambient light value due to BCP ($A_{BCP}$), which is then combined with the ambient light value due to DCP ($A_{DCP}$) in a weighted fashion $A' = \lambda A_{DCP} + (1-\lambda) A_{DCP}$. Similarly, the transmission maps due to the BCP and DCP are computed in the same manner as $t'(x) = \Lambda t_{BCP}(x) + (1-\Lambda) t_{DCP}(x)$, and the final $A'$ is used to recover the dehazed image.

The density classification prior was introduced by Yang et.al. \cite{yang2023single}, also in 2023. This is based on the observation that haze enhances the brightness of the image, and computing the channel differences allows determination of the haze densities. This is mathematically illustrated as follows:

\begin{equation}
    f(x) = I^{c}_{max}(x) - I^{c}_{min}(x),
\end{equation}

where $I^{c}_{max}(x) = \max_{c \in (r,g,b)}\left(I^{c} (x) \right)$, $I^{c}_{min}(x) = \min_{c \in (r,g,b)}\left(I^{c} (x) \right)$. Unlike a typical low to moderate haze image, a dense haze image has $f(x) \rightarrow 0$ as its contrast is severely degraded; thus there are only minute changes in the channel differences. Therefore their method is able to easily distinguish between low (or moderate) haze and dense haze images via atmospheric light veil estimation of the form 

\begin{equation}
    V(x) = I^{c}_{min}(x)^{\gamma(x)},
\end{equation}

where $\gamma(x)$ is the automatic adjustment function parameter. For the dense haze scenario, the atmospheric light veil value was greater, and vice versa for low to moderate haze. Therefore $\gamma(x)$ is related to $f(x)$ by $\gamma(x) = mean(1-f(x))$. The $V(x)$ can then be used to recover the haze-free images.

Zhao \cite{zhao2021single} introduced a bounded channel differences (BCD) prior in 2021 that utilized a proper bounding function to estimate the local transmission map more accurately by analyzing the prior information of the difference maps across the RGB channels of the dehazed image patches using ambient light values. This is based on formulating a metric function $M_{0}$ as the summation of the absolute values of the dehazed patch map across the RGB channels:

\begin{equation}
    M_{0} = \left|\left(\frac{I_{r}}{A_{r}} - \frac{I_{g}}{A_{g}} \right) \frac{1}{t} \right|_{1} + \left| \left(\frac{I_{r}}{A_{r}} - \frac{I_{b}}{A_{b}} \right) \frac{1}{t} \right|_{1} + \left| \left(\frac{I_{b}}{A_{b}} - \frac{I_{g}}{A_{g}} \right) \frac{1}{t} \right|_{1},
\end{equation}

where $I$ and $A$ denotes the dehazed images and ambient light in the color channels, respectively. An issue arose in extreme color cases where $J_{r}(x) = 1, J_{g}(x) = 1, J_{b}(x) = 0$, (where $J$ denotes the hazy images) which could lead to over-dehazing in the original hazy patches, thereby losing the relevant contrasts at these portions. To mitigate the challenge, the bounding function $R(J)$ is proposed that acts as a regularization term (i.e., $M = M_{0} + R(J)$), where

\begin{equation}
    R(J) = \textbf{F}\left(\frac{J_{c1}}{J_{c2}}\right),
\end{equation}

subjected to the condition $c_{1,2} \in (r,g,b), c_{1} \neq c_{2}$. $\bold{F}$ is a checking function that ensures that $M \rightarrow \pm \infty$ when a value of 0 exists in the dehazed patches. The final transmission map can then be estimated by maximizing the bounded metric function of

\begin{equation}
    \tilde{t} = \max_{t}M_{0} + R(J).
\end{equation}

where $t \in (\epsilon,1) $ and $\epsilon$ is a small positive threshold.

Table 3 summarizes the prior-based dehazing approaches covered thus far, with the type of estimation parameters, year in which the method was published, and some of the datasets in which each method was evaluated.

\begin{table*} 
\centering
\caption{Summary of prior-based dehazing as described, including the year it was published, the relevant dehazing parameters, and some of the datasets it was evaluated on.}
\begin{tabular}{p{3cm}p{1cm}p{3cm}p{5.5cm}}
\hline
\textbf{Priors} & \textbf{Year} & \textbf{Estimation parameters} & \textbf{Some Datasets Evaluated} \\
\hline
SSP \cite{fattal2008single} & 2008 & $t(x), A$ & Custom images from internet\\
\hline
DCP \cite{he2010single} & 2009 & $t(x)$, $A$ & Custom images from internet, RESIDE \cite{li2018benchmarking}, NH-HAZE \cite{ancuti2020nh}, DENSE-HAZE \cite{ancuti2019dense}, O-HAZE \cite{ancuti2018haze}, I-HAZE \cite{ancuti2018hazeI} \\
\hline
BCCR \cite{meng2013efficient} & 2013 & $t(x)$, $A$ & Custom images from internet, RESIDE \cite{li2018benchmarking}, NH-HAZE \cite{ancuti2020nh}\\
\hline
OCE \cite{kim2013optimized} & 2013 & $t(x)$, $A$ & Custom images from internet \\
\hline
CAP \cite{zhu2015fast} & 2015 & $d(x)$ & Custom images from internet, RESIDE \cite{li2018benchmarking}, NH-HAZE \cite{ancuti2020nh}  \\
\hline
NLD \cite{berman2016non} & 2016 & $t(x)$, $A$ & Custom images from internet, RESIDE \cite{li2018benchmarking}\\
\hline
BCD \cite{zhao2021single} & 2021 & $\tilde{t}$, $M$ &  Custom images from internet, RESIDE \cite{li2018benchmarking}\\
\hline
BCP+DCP \cite{li2023single} & 2023 & $t'(x)$, $A'$ & RESIDE \cite{li2018benchmarking}, O-HAZE \cite{ancuti2018haze}, NH-HAZE \cite{ancuti2020nh}\\
\hline
Density Class \cite{yang2023single} & 2023 & V(x) & RESIDE \cite{li2018benchmarking}, O-HAZE \cite{ancuti2018haze}, I-HAZE \cite{ancuti2018hazeI}, NH-HAZE \cite{ancuti2020nh}\\\\
\hline
\end{tabular}
\label{tab:table3}
\end{table*}

%Patch size, radius, regularization factor
%Patch size, bounded parameter, regularization factor 
%Patch size, bounded parameter, regularization factor 
%Patch size, radius, regularization factor
% Radius, $\theta_{0}(x),\theta_{1}(x),\theta_{2}(x)$ 
% Radiometric correction parameter 
% $R(J)$, regularization factor
% Patch size, radius, regularization factor
% $\gamma(x)$, radius

\subsection{Deep learning approaches for image dehazing}

As mentioned previously, in our review, learning-based dehazing approaches can be categorized into CNN-based dehazing, ViT-based dehazing, and contrastive and few-shot dehazing.

\subsubsection {CNN-based image dehazing}

Cai et.al. \cite{cai2016dehazenet} presented DehazeNet, arguably one of the first CNN-based end-to-end dehazing network that required only hazy-clear corresponding image pairs under the guidance of the ASM. Their network introduced the Bilateral Rectifier Linear Unit (BReLU), based on the authors' observation that the ReLU may not be a good activation function for regression problems (such as dehazing). The BReLU maintains bilateral restraint and local linearity, which mitigates the response overflow problem for the last layer in an image restoration problem, for which the output value in the last layer is to be bounded at both the upper and lower limits at small ranges. Their network comprised only four CNN layers, where the network parameters to be learned included the filters and biases of each layer.

Li et.al. \cite{li2017aod} introduced the All-in-One Dehazing Network (AOD-Net), which is one of the first lightweight CNN incorporated with ASM. The AOD-Net utilizes a modified ASM formula, which is described by 

\begin{equation}
    J(x) = K(x)I(x) - K(x) + b,
\end{equation}

where $K(x)$ represents the following term, which unifies $t(x)$ and $A$ into one formula:

\begin{equation}
    K(x) = \frac{\frac{1}{t(x)}(I(x) - A) + (A + b)}{I(x) - 1}.
\end{equation}

The AOD-Net comprises nine layers, with five convolutional layers utilized to estimate $K(x)$, and a clean image generation module with an element-wise multiplication layer and multiple element-wise addition layers to recover the haze-free image by applying Equation 19.

Ren et.al. \cite{ren2020single} introduces a Multi-Scale CNN (MSCNN) for single image dehazing that utilizes a coarse detail network that estimates the transmission map of the hazy images holistically, and a fine detail network that refines the dehazed outputs from the coarse network in a local manner. Both the coarse network and the fine-grained network are comprised of a convolutional layer, a max pooling layer, and an upsampling layer arranged in a sequential order. There are three sets of such combinations in both networks before the respective intermediate outputs are fed into a linear combination layer, before being fed into the ASM to recover the dehazed images. To the best of our knowledge, this is one of the first dehazing works to emphasize detail recovery by incorporating fine detail architecture, which is commonly utilized in related image deraining works, but with little focus in the dehazing literature.

Qin et al. \cite{qin2020ffa} proposed a feature-fusion attention network (FFA-Net), which includes an attention-based feature fusion method. It integrates both channel and spatial (pixel) attention to address the various features in an uneven manner and increase the CNN's representational capabilities and adaptability to handle various kinds of data. In the channel-wise network, global average pooling is used to process global spatial information to obtain the channel descriptors $g_{c}$ as follows:

\begin{equation}
    g_{c} = GlobalPooling(\phi_{c}) = \frac{1}{HW}\sum_{i=1}^{H}\sum_{i=1}^{W} X_{c} (i,j),
\end{equation}

where $X_{c} (i,j)$ represents the value of channel $X_{c}$ at the pixel position $(i,j)$, and $H$ and $W$ denote the height and width of the image, respectively. The ReLU and sigmoid $\sigma$ activation functions were then applied to the descriptors to obtain the feature weights of the different channels.

\begin{equation}
    Attention_{c} =  \sigma(conv(ReLU(conv(g_{c})))).
\end{equation}

The modified feature maps from the channel attention are obtained via element-wise multiplication with the input feature maps as $\phi'_{c}$ as $\phi_{c} = Attention_{c} \otimes \phi_{c}$. In the pixel-wise network, $\phi'_{c}$ is fed into two convolutional layers with the previously described activations to obtain $Attention_{p}$ as 

\begin{equation}
    Attention_{p} =  \sigma(conv(ReLU(conv(\phi'_{c})))),
\end{equation}

and similar to that of channel-wise attention, the resultant feature output $\phi''_{p}$ is obtained by $\phi''_{p} = Attention_{p} \otimes \phi'_{c}$. Their proposed network also introduced a basic block comprising of local residual learning, which enables the information gained from the thin-haze regime, or low-frequency information, to be skipped via multiple skip-connection layers.

Liu et al. \cite{liu2019griddehazenet} presented GridDehazeNet, a multi-scale estimation on a grid network that reduces the bottleneck problem frequently presented in conventional multi-scale techniques via an attention mechanism. Unlike the other techniques described so far, GridDehazeNet does not rely on  ASM. Their network comprises a pre-processing module, a backbone module, and a post-processing module. The pre-processing block comprised a residual dense block (RDB) as well as a convolutional layer without an activation function. From the hazy images, it creates 16 feature maps, which are referred to as the learned inputs. The backbone module utilizes the improved GridNet \cite{fourure2017residual}, which was first implemented for semantic segmentation. Using the learned inputs provided by the pre-processing module, an attention-based multi-scale estimation is implemented. Unlike the pre-processing module which comprise of only the RDB, the backbone includes the downsampling and upsampling modules, along with attention-based feature addition. Finally, the post-processing module enhances the quality of the dehazed images' via artifact removal, and its algorithmic architecture is symmetrical to that of the pre-processing module. The latter module works in a manner analogous to the fine-detail network in the MSCNN.

A pyramidal architecture was introduced by Zhang and Patel \cite{zhang2018densely} and named the Densely Connected Pyramid Dehazing Network (DCPDN). The network can simultaneously learn the transmission map, the airlight and the dehazing task, and incorporated a densely connected encoder-decoder structure to maximize information flow and leverage features from different layers in estimating the transmission map. Because the structure of the map and the dehazed images were highly correlated, a GAN-based joint discriminator was utilized to determine whether the paired samples (of the map and dehazed image) were from the same data distribution. ASM was also incorporated in this study. 

By considering complementary information in the encoder and decoder, as well as a bottleneck residual block layer that modifies the weights of the local gradients at various scales, Yi et al. \cite{yi2021msnet} proposed a multiple inter-scale dense skip-connection dehazing network (MSNet) that extends the traditional U-Net architecture. The network also incorporates an inter-scale feature reuse mechanism that allows better learning of the intrinsic relationship between $I(x)$ and $J(x)$ in an end-to-end manner. In the encoder stage, there were multiple downsample skip connections, whereas in the decoder stage, there were multiple upsample skip connections. Their network did not incorporate the ASM.

There are methods that utilized Knowledge Distillation (KD), which entailed a teacher-student model in which the teacher model transferred its knowledge or weights to the student model to ensure optimal performances at higher computational efficiency, or under smaller sample size. Hong et.al. \cite{hong2020distilling} proposed a KD Dehazing Network (KDDN) that distills networks with heterogeneous task imitation. The teacher network featured an off-the-shelf autoencoder network for image reconstruction, and the student network imitated the teacher network's tasks. A spatially-weighted residual channel attention block was also devised for the student network to learn contextually aware channel attention in an adaptive manner. No atmospheric scattering model was used in this proposed network. An Online Knowledge Distillation Network (OKDNet) was proposed by Lan et.al. \cite{lan2022online} which pre-processed the input hazy images before utilizing a multi-scale network comprising attention-based residual dense blocks to obtain its shared features. The intermediate features are then inputted into different branches. The first branch is used to estimate the dehazing images via ASM, and the other branch relates the hazy and ground-truth images in a model-free manner. Finally, an aggregation block is utilized to fuse the information from the two branches efficiently, and the joint optimization performs an online one-stage KD. A Haze-Aware Representation Distillation GAN (HardGAN) was introduced by Deng et.al. \cite{deng2020hardgan} which implements the Haze-Aware Representation Distillation (HARD) module to not only include the normalization layer, but also aggregate spatial information and ambient light caused by the different levels of haze intensity attentively using the haze-aware map. The GAN network was implemented to learn style transfer mapping directly.

All the aforementioned methods performed well for homogeneous haze but may not fare well for inhomogeneous haze (as in NH-HAZE \cite{ancuti2020nh}) and thick haze (as in DENSE-HAZE \cite{ancuti2019dense}). This is because a single neural network dehazing model is not capable of modelling different haze intensities. Ensemble learning has been shown to reduce variances in neural networks, implying that the ensemble model can perform better than the best single neural network model. In view of this, Yu et.al. \cite{yu2020ensemble} first introduced EDN-3J, which comprises one shared dense encoder and four dense decoder modules. Three of the decoders output a distinct $\textbf{J}$ value obtained using different reconstruction loss functions in a weighted manner. The final decoder effectively generated weight maps for combining the three decoder outputs. Second, they introduced EDN-AT, which incorporated the ASM at the end, which also included the same configurations as those of EDN-3J, but with two decoders outputting the ambient light maps and one decoder outputting the atmospheric transmission map. Finally, they also proposed EDN-EDU, which consists of a sequential ensemble of dense encoder, decoder, and U-Net. The purpose in this case is to cascade the encoder-decoder network with different capabilities such that they generalize well to complex haze scenario. A two-branch CNN-based neural network was introduced by Yu et.al. \cite{yu2021two}, for which the first branch is the transfer learning network already pretrained using ImageNet. The purpose was to alleviate over-fitting owing to the lack of large-scale inhomogeneous haze datasets, and to provide robust transfer learning features. The second branch involved a current data-fitting sub-net built upon via the Residual Channel Attention Network (RCAN), which maintained the original resolution of the inputs by not including any downsampling module, thus retaining the finer-detail features of the inputs. A fusion tail is included at the end of the ensemble strategy. No ASM was included in this study.

A Deep hybrid network was introduced by Jiang et.al. \cite{jiang2023deep}, which once again considers detail refinement as in MSCNN along with dehazing. For dehazing, a squeeze and excite (SE) block \cite{hu2018squeeze} was utilized and constitute their haze residual attention sub-network branch. The detail refinement sub-network branch takes as input the dehazed images and utilizes the multi-scale contextual information aggregation. By training both branches in a joint manner, effective dehazing can be achieved without compromising the details and introducing unwanted artifacts.

There has also been a recent emergence in utilizing Graph Convolutional Network (GCN) for dehazing. For instance, Hu et.al. \cite{hu2023hazy} proposed a network that combines CNN and GCN to capture both local and global haze features. The triple attention module was incorporated into the CNN structure to emphasize more weights for local features, whereas the dual GCN module was incorporated to fuse spatial coherence with channel correlation to extract global features. Their overall architecture was inspired by U-Net.

Table 4 summarizes the CNN-based dehazing approaches covered thus far, along with the year in which the work was published, whether the ASM was incorporated, and some of the datasets in which the work was evaluated.

\begin{table*} 
\caption{Summary of CNN-based dehazing as described, including the year it was published, whether if ASM is incorporated, and some of the datasets it was applied on.}
\begin{tabular}{p{3cm}p{1cm}p{2cm}p{6cm}}
\hline
\textbf{Works} & \textbf{Year} & \textbf{ASM included?} & \textbf{Some Datasets Evaluated} \\
\hline
DehazeNet \cite{cai2016dehazenet} & 2016 & Yes &  Middlebury stereo images \cite{scharstein2002taxonomy}, RESIDE \cite{li2018benchmarking},  NH-HAZE \cite{ancuti2020nh}, O-HAZE \cite{ancuti2018haze}, I-HAZE \cite{ancuti2018hazeI}, DENSE-HAZE \cite{ancuti2019dense}\\
\hline
AOD-Net \cite{li2017aod} & 2017 & Yes & NYU-Depth2 \cite{silberman2012indoor}, RESIDE \cite{li2018benchmarking}, O-HAZE \cite{ancuti2018haze}, I-HAZE \cite{ancuti2018hazeI}, NH-HAZE \cite{ancuti2020nh}, DENSE-HAZE \cite{ancuti2019dense} \\
\hline
DCPDN \cite{zhang2018densely} & 2018 & Yes & NYU-Depth2 \cite{silberman2012indoor}, RESIDE \cite{li2018benchmarking},  NH-HAZE \cite{ancuti2020nh}, I-HAZE \cite{ancuti2018hazeI}, O-HAZE \cite{ancuti2018haze}\\
\hline
GridDehazeNet \cite{liu2019griddehazenet} & 2019 & No & RESIDE \cite{li2018benchmarking}, NH-HAZE \cite{ancuti2020nh}, DENSE-HAZE \cite{ancuti2019dense}, O-HAZE \cite{ancuti2018haze}, I-HAZE \cite{ancuti2018hazeI}\\
\hline
MSCNN \cite{ren2020single} & 2020 & Yes & NYU-Depth \cite{silberman2012indoor}, RESIDE \cite{li2018benchmarking}, I-HAZE \cite{ancuti2018haze}, O-HAZE \cite{ancuti2018haze}\\
\hline
FFA-Net \cite{qin2020ffa} & 2020 & Yes & RESIDE \cite{li2018benchmarking}, NH-HAZE \cite{ancuti2020nh}, O-HAZE \cite{ancuti2018haze}, I-HAZE \cite{ancuti2018hazeI}, DENSE-HAZE \cite{ancuti2019dense}\\
\hline
KDDN \cite{hong2020distilling} & 2020 & No & RESIDE \cite{li2018benchmarking}, O-HAZE\cite{ancuti2018haze}, NH-HAZE \cite{ancuti2020nh}, DENSE-HAZE \cite{ancuti2019dense} \\
\hline
HardGAN \cite{deng2020hardgan} & 2020 & No & RESIDE \cite{li2018benchmarking}, NH-HAZE\cite{ancuti2020nh} \\
\hline
EDN-3J \cite{yu2020ensemble} & 2020 & No & NH-HAZE\cite{ancuti2020nh}, DENSE-HAZE \cite{ancuti2019dense}\\
\hline
EDN-AT \cite{yu2020ensemble} & 2020 & Yes & NH-HAZE\cite{ancuti2020nh}, DENSE-HAZE \cite{ancuti2019dense}\\
\hline
EDN-EDU \cite{yu2020ensemble} & 2020 & No &  NH-HAZE\cite{ancuti2020nh}, DENSE-HAZE \cite{ancuti2019dense}\\
\hline
MSNet \cite{yi2021msnet} & 2021 & No & RESIDE \cite{li2018benchmarking} \\
\hline
2-branch \cite{yu2021two} & 2021 & No & RESIDE \cite{li2018benchmarking}, O-HAZE \cite{ancuti2018haze}, DENSE-HAZE \cite{ancuti2019dense}, NH-HAZE \cite{ancuti2020nh}, NTIRE 2021 \cite{ancuti2021ntire} \\
\hline
OKDNet \cite{lan2022online} & 2022 & Yes & RESIDE \cite{li2018benchmarking}, I-HAZE \cite{ancuti2018haze}, O-HAZE \cite{ancuti2018hazeI} \\
\hline
DHN \cite{jiang2023deep} & 2023 & No & RESIDE \cite{li2018benchmarking}, NH-HAZE \cite{ancuti2020nh}\\
\hline
GCN \cite{hu2023hazy} & 2023 & No & RESIDE \cite{li2018benchmarking}, DENSE-HAZE \cite{ancuti2019dense}, NH-HAZE \cite{ancuti2020nh} \\
\hline
\end{tabular}
\label{tab:table4}
\end{table*}

\subsubsection{ViTs-based image dehazing}

One of the first ViT-based dehazing methods, the DehazeFormer, was introduced by Song et.al. \cite{song2023vision}, which utilizes the swin transformer architecture \cite{liu2021swin} along with a modified normalization layer (which utilized the Rescaled Normalization instead of the usual Layer Normalization), a spatial information aggregation scheme, and a modified activation function (softRELU instead of the GELU and ReLU). These modifications were made as the authors observed that using the typical normalization layer and activation function decreased the ViT performance in the dehazing task. They tested different variants of the DehazeFormer, specifically  DehazeFormer-T, DehazeFormer-S, DehazeFormer-B, DehazeFormer-M and DehazeFormer-L, and found that the DehazeFormer-M and DehazeFormer-L have the best performance on all the datasets evaluated. Furthermore, they also introduced the RS-HAZE which consists of high-level clouds and haze removal from a remote sensing perspective.

The DHFormer was introduced by Wasi and Shiney \cite{wasi2023dhformer} which utilized residual learning and ViT in an attention module, and comprised two networks. The first network uses the ratio of the hazy image $J(x)$ and the transmission matrix to arrive at the residual map. The second network utilizes the map, passes it through multiple convolutional layers, and superpose the intermediate outputs into the generated resultant feature maps. Finally, a global-context and depth-aware transformer encoder is utilized to obtain channel attention, which can be used to deduce the spatial attention map to arrive at the final dehazed images.

Guo et al. \cite{guo2022image} proposed Dehamer, which was one of the first works to combine a CNN with ViT for dehazing. They suggested a feature modulation technique on CNN-captured features to address feature inconsistency between the CNN and transformer modules. This allows their procedure to simultaneously inherit the global and local feature modeling capabilities of the CNN and transformer respectively. Zhao et al. \cite{zhao2021complementary} presented a Hybrid Local-Global (HyLoG) attention-based architecture that combined local and global transformer paths to simultaneously capture global and local feature dependencies in pictures. Complementary Feature Selection Module (CFSM) was incorporated to adaptively select crucial complementary feature for dehazing tasks. A Swin Transformer U-Net architecture (ST-UNet) was introduced by Bian et.al. \cite{bian2022swin} for effective dehazing while significantly increasing the processing time. Parallel processing and data structure modification were also utilized in the prediction workflow to maximize computational resources. This is one of the few dehazing works that has been applied to dehazing images for post-earthquake scene assessment and analysis, from a remote sensing perspective.

The Trinity-Net introduced by Chi et.al. \cite{chi2023trinity} combined prior and deep learning-based strategies to restore realistic surface information. A gradient-guidance module that utilizes structure priors from gradient maps allows the generation of visually pleasing dehazing outputs. Both CNN and ViT architectures were utilized to interpret prior information and arrive at a reasonable haze parameter estimation. The work was also motivated by the scarcity of data available in the military domain, which prompted them to develop a Remote Sensing Image Dehazing benchmark (RSID) along with a Natural Image Dehazing (NID) benchmark to evaluate their work against SOTAs. A Deep Guided Transformer Dehazing Network (DGDTN) was introduced by Zhang et.al. \cite{zhang2023deep} which combined and incorporated a CNN to capture local information, with the transformer simultaneously capturing long-range dependency. A guided filter is also utilized to accelerate the dehazing processing time. A ViTGAN network that combines a vision transformer and generative adversarial network was proposed by Liu \cite{liu2023image}. Both the discriminator and generator networks used the ViT architecture; however, because the training would be unstable, a generator structure optimization along with discriminator regularization was utilized. A two-stage dehazing network incorporating a swin transformer was proposed by Li et.al. \cite{li2022two}, which includes an inter-block supervision mechanism between the encoder and decoder modules for feature refinement, supervision and selection, thus enhancing transmission efficiency. In addition, a fusion attention mechanism is added between the different stages of the network architecture to ensure the transmission authenticity of the feature signals in the first stage and further refine the efficiency of the model.  

A transformer-based wavelet network for dehazing (WaveletFormerNet) was proposed by Zhang et.al. \cite{zhang2024waveletformernet} which incorporated the Discrete Wavelet Transform (DWT) into the ViT and introduced the WaveletFormer and IWaveletFormer blocks to reduce texture detail loss, color artifacts and distortion owing to downsampling. Parallel convolution is also included in the transformer blocks which allows the capturing of multiple information about the image frequencies without overstacking the convolutional blocks, thereby increasing the efficiency of the model. The Feature Aggregation Module (FAM) was also utilized to enhance the feature extraction function while retaining the resolution.

Similar to CNN-based approaches, ensemble learning has also been utilized for transformer architectures. For example, TransER was introduced by Hoang et.al. \cite{hoang2023transer} which comprises the TransConv fusion dehaze (TFD) model in stage one and lightweight ensemble reconstruction in stage two. The TFD model is an ensemble-based architecture consisting of one shared ViT-based encoder with a layout inspired by the TransConv module along with three separate corresponding decoders with skip connections similar to U-Net. Numerous attention channels from the feature maps of the encoder and decoder are linked using Selective Kernel Fusion (SKF) \cite{song2022rethinking}, which has been shown to be more efficient than the typical concatenation. Knowledge distillation (KD) is implemented in stage two to distill learned weights from the teacher network to the student network to better leverage the intermediate feature maps and subsequently the distribution of the haze-free images from the TRN. Finally, a Lightweight Ensemble Reconstruction (LER) was introduced as a ensemble reconstruction network that comprised of two encoders and one encoder, and make use of the Gated Convolution Module (GCM) for the feature maps extraction and combination in an adaptive manner to maximize the preservation of information.

Table 5 summarizes the ViT-based dehazing approaches covered above, along with the year in which the work was published, whether ASM was incorporated, and some of the datasets in which the work was evaluated. One immediate trend we can observe is that ViT-based dehazing has started to gain attention and be incorporated more recently (post 2021), with increasing work in this direction in 2023, as shown in the same table.

\begin{table*} 
\caption{Summary of ViT-based dehazing as described, including the year it was published, whether if ASM is incorporated, and some datasets it was applied on.}
\begin{tabular}{p{4cm}p{1cm}p{2cm}p{5.5cm}}
\hline
\textbf{Works} & \textbf{Year} & \textbf{ASM included?} & \textbf{Some Datasets Evaluated} \\
\hline
HyLoG \cite{zhao2021complementary} & 2021 & No & RESIDE \cite{li2018benchmarking}, NH-HAZE \cite{ancuti2020nh}, O-HAZE \cite{ancuti2018haze}, NHR \cite{zhang2020nighttime} \\
\hline
DehazeFormer \cite{song2023vision} & 2022 & No & RESIDE \cite{li2018benchmarking}, RS-HAZE \\
\hline
Dehamer \cite{guo2022image} & 2022 & No & RESIDE \cite{li2018benchmarking}, NH-HAZE \cite{ancuti2020nh}, O-HAZE \cite{ancuti2018haze}, I-HAZE \cite{ancuti2018hazeI}, DENSE-HAZE \cite{ancuti2019dense}\\
\hline
ST-UNet \cite{bian2022swin} & 2022 & No & AID30 \cite{xia2017aid}, RSSCN8 \cite{zou2015deep} \\
\hline
2-stage swin ViT \cite{li2022two} & 2022 & No & RESIDE \cite{li2018benchmarking}, I-HAZE \cite{ancuti2018haze}, O-HAZE \cite{ancuti2018hazeI} \\
\hline
DHFormer \cite{wasi2023dhformer} & 2023 & Yes & RESIDE \cite{li2018benchmarking} \\
\hline
Trinity-Net \cite{chi2023trinity} & 2023 & Yes & RSID, NID (self-made)\\
\hline
DGDTN \cite{zhang2023deep} & 2023 & No & RESIDE \cite{li2018benchmarking}, NH-HAZE\cite{ancuti2020nh}  \\
\hline
TransER \cite{hoang2023transer} & 2023 & Yes & RESIDE \cite{li2018benchmarking}, NH-HAZE\cite{ancuti2020nh}, NTIRE 2021 \cite{ancuti2021ntire}, NTIRE 2023 \cite{li2023ntire} \\
\hline
ViTGAN \cite{liu2023image} & 2023 & No & Datasets from \cite{engin2018cycle} \\
\hline
WaveletFormerNet \cite{zhang2024waveletformernet} & 2024 & No & RESIDE \cite{li2018benchmarking}, I-HAZE \cite{ancuti2018haze}, O-HAZE \cite{ancuti2018hazeI}, NH-HAZE \cite{ancuti2020nh}, DENSE-HAZE \cite{ancuti2019dense}\\
\hline
\end{tabular}
\label{tab:table5}
\end{table*}

\subsubsection{Contrastive learning for image dehazing}

The AECR-Net proposed by Wu et.al. \cite{wu2021contrastive} combined an autoencoder-like architecture  with contrastive regularization, which is inspired by the triplet loss network. In their work, the anchor images were the dehazed images, the positive images were the ground-truth clear images, and the negative images were the original hazy images. The loss function is devised such that the embedded anchor and positive image features are as close to each other as possible, whereas the embedded anchor and negative image features are as far apart as possible. Mathematically, this is manifested as the sum of the reconstruction loss and regularization terms, written as follows:

\begin{equation}
    \mathcal{L}_{rec} + \mathcal{L}_{contrastive} =  ||J(x) - I(x) ||_{1} + \mathcal{B} \sum_{i=1}^{N} \Omega_{i} \frac{||G_{i}(J) - G_{i}(I(x)) ||_{1}}{||G_{i}(J) - G_{i}(I(x))||_{1}},
\end{equation}

where $\mathcal{B}$ and $\Omega_{i}$ denote the two hyperparameters for balancing the weights of the respective losses during training and $G_{i}(\cdot), i = 1,2,...N$ represents the i$\textsuperscript{th}$ hidden features from the respective feature extraction architecture. 

Using Contrastive Disentanglement Learning (CDL), Chen et.al. \cite{chen2022unpaired} proposed an unpaired learning-based dehazing paradigm that treats the problem as a two-class separated factor disentangled task. A cycleGAN framework is proposed to guide the learning of disentangled representations by associating the generated images with latent factors. A generator is used to generate negative adversarial samples, which are then trained in an end-to-end manner with the representative network to better obtain the information required for discrimination, thereby improving the disentanglement factor by optimizing the adversarial contrastive loss. Their experiment also demonstrated that hard-negative samples can suppress task-irrelevant factors, whereas unpaired haze-free samples can improve task-relevant factors, which collectively improve dehazing performances.

A Hierarchical Contrastive Dehazing (HCD) paradigm was proposed by Wang et.al \cite{wang2024restoring} utilizing feature fusion and contrastive learning. The hierarchical interaction module utilizes multi-scale activation to exploit the hierarchy of the features in the network and revise their responses. A novel hierarchical loss function that performs the contrastive learning on paired exemplars guides the image dehazing. A low-frequency sub-band contrastive regularization (LSCR) module was incorporated in the dehazing task by Bai and Yuan \cite{bai2022contrastive} using the wavelet domain. Their motivation was that existing dehazing approaches neglect the fact that the appearance of haze mainly affect the low-frequency components of the image, meaning that the different components are processed indiscriminately, resulting in lower dehazing results. By utilizing these approaches, the components of the anchor images that are greatly affected by haze are pulled closer to the ground-truth images and pushed further away from the hazy images. A high-frequency sub-band loss was also introduced concurrently to ensure that the higher-frequency components of the dehazed images were consistent with the ground-truth images. 

Zheng et.al. \cite{zheng2023curricular} introduced Curricular Contrastive Regularization for physics-aware dehazing that attempts to make the contrastive dehazing tasks consensual (i.e., increasing the diversity of the negative samples by mixing the original hazy images with the dehazed images obtained from other methods). This mitigates the under-constrained solution space problem in that only the original hazy images serve as lower-bound constraints. More specifically, the negative samples were categorized into three levels of difficulty: easy (E), hard (H), and ultra-hard (U), in which the original hazy images were classified as E, and a coarse strategy was used to differentiate the other two cases. Different weights are then assigned to the aforementioned negative pairs, and a physics-aware dual-branch unit (PDU) is also implemented that approximates ambient light features and transmission map in dual branches using physics-based priors, hence synthesizing the features of the latent clear images in a more precise manner than solely using the ASM. Their C$^{2}$ PNet implemented the PDUs in a cascaded backbone manner. 

\subsubsection{Few-shot dehazing}

The zero-shot image dehazing (ZID) paradigm was introduced by Li et.al. \cite{li2020zero} which made use of the assumption of layer disentanglement and entailed viewing an image as entangled by many simple layers. This means that for hazy images, one can view the haze-free images, transmission maps, and ambient lights as three layers that can be entangled to form haze images. The proposed ZID utilizes three joint sub-networks, J-Net, T-Net, and A-Net, to disentangle the input into the three layers and subsequently estimate the haze parameters and recover the haze-free image. Their approach is a special type of zero-shot method in that it exploits only the information in the observed single hazy image and does not require training on datasets. This contrasts with typical zero-shot learning in other visual tasks, in that they require a trained model to infer unseen categories. Hence, ZID can avert laborious data collection and also avoid the the domain shift issue due to the use of synthetic hazy images to tackle real-world hazy scenarios.

A zero-shot network that estimates the ASM parameters to perform image dehazing was proposed by Kar et.al. \cite{kar2021zero}. This is based on their observation that a suitable deterioration of the hazy image implies a controlled perturbation of the ASM parameters that describes the relationship between the original unperturbed image and the input image. Zero-shot learning is achieved via an optimization procedure that maintain the aforementioned relationship between two pairs of ASM parameters obtained from the input and its deteriorated form. Their method was also evaluated using real-world underwater images displaying varying levels of contrast and color. A related study that combined the layer disentanglement paradigm with that of their novel re-degradation haze model for zero-shot dehazing was proposed by Wei et.al. \cite{wei2022zero} and evaluated their approach using remote sensing imagery.

Table 6 summarizes the contrastive and few-shot-based dehazing approaches discussed thus far, along with the year in which the work was published, whether ASM was incorporated, and some of the datasets in which the work was evaluated.

\begin{table*} 
\caption{Summary of contrastive and few-shot-based dehazing as described, including the year it was published, whether if ASM is incorporated, and some of the datasets it was applied on.}
\begin{tabular}{p{4cm}p{1cm}p{2cm}p{5.5cm}}
\hline
\textbf{Works} & \textbf{Year} & \textbf{ASM included?} & \textbf{Some Datasets Evaluated} \\
\hline
ZID \cite{li2020zero} & 2020 & Yes & RESIDE \cite{li2018benchmarking}, O-HAZE \cite{ancuti2018haze}\\
\hline
ZS perturbation \cite{kar2021zero} & 2021 & Yes & I-HAZE \cite{ancuti2018hazeI}, O-HAZE \cite{ancuti2018haze}\\
\hline 
AECR-Net \cite{wu2021contrastive} & 2021 & No & RESIDE \cite{li2018benchmarking}, NH-HAZE \cite{ancuti2020nh}, DENSE-HAZE \cite{ancuti2019dense}, O-HAZE \cite{ancuti2018haze}, I-HAZE \cite{ancuti2018hazeI}\\
\hline
Wei et.al. \cite{wei2022zero} & 2022 & Yes & RESIDE \cite{li2018benchmarking}, RS-HAZE, RICE-I \cite{lin2019remote}, DeepGlobe \cite{demir2018deepglobe}\\
\hline
CDL \cite{chen2022unpaired} & 2022 & No & RESIDE \cite{li2018benchmarking}, Foggy CityScapes \cite{sakaridis2018semantic} \\
\hline
LSCR \cite{bai2022contrastive} & 2022 & No & RESIDE \cite{li2018benchmarking}, DENSE-HAZE \cite{ancuti2019dense}\\
\hline
C$^{2}$PNet \cite{zheng2023curricular} & 2023 & Yes & RESIDE \cite{li2018benchmarking}, I-HAZE \cite{ancuti2018hazeI}, NH-HAZE \cite{ancuti2020nh}, DENSE-HAZE \cite{ancuti2019dense}, NTIRE 2021 \cite{ancuti2021ntire}\\
\hline
HCD \cite{wang2024restoring} & 2024 & No & RESIDE \cite{li2018benchmarking}, HazeRD \cite{zhang2017hazerd}, DENSE-HAZE \cite{ancuti2019dense}\\
\hline
\end{tabular}
\label{tab:table6}
\end{table*}

% Put this section before the Results and Discussions section. Thanks.

\section{Comparative analysis of image dehazing techniques}

\begin{figure}[hbt!]
    \centering
    \includegraphics[scale=0.80]{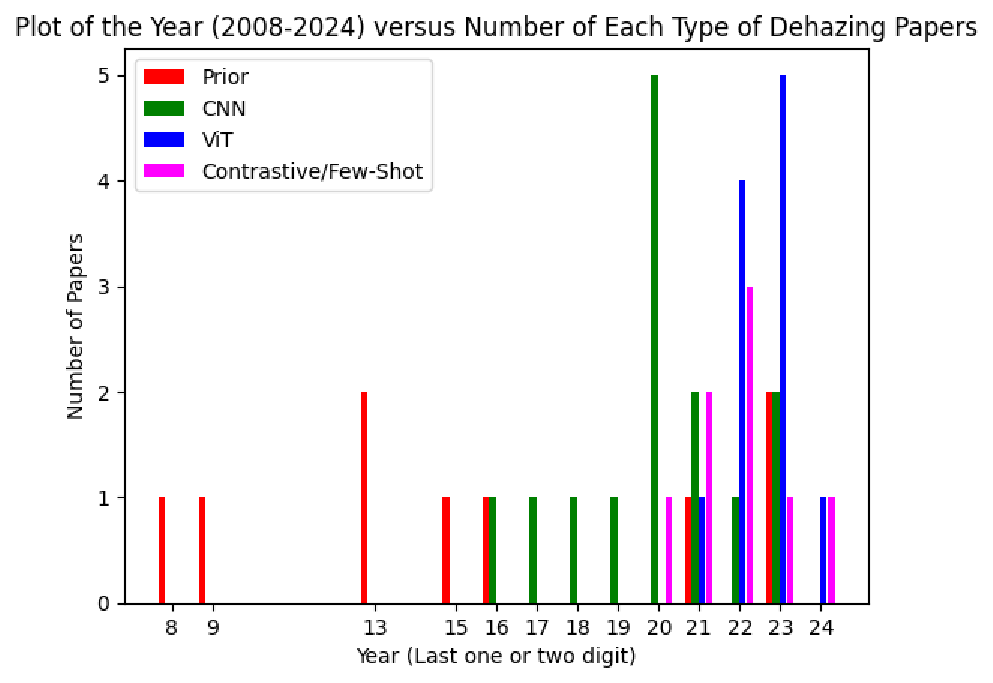}
    \caption{Bar graphs of frequency of each type of dehazing approach against the year (last one or two digit) in which it was published, based on all the works described so far. Bars in red denote prior-based works, bars in green denote CNN-based works, bars in blue denote ViT-based works, and bars in magenta denote contrastive or few-shot based works.}
    \label{Bar_Plots}
\end{figure}

Figure \ref{Bar_Plots} shows the bar graphs of the frequency of each type of dehazing approach against the year it was published, based on all the works described so far. The years range from 2008 to 2024. We can immediately observe that CNN-based dehazing started to be focused on in 2016 (before that, prior-based dehazing works dominated), and received the most attention based on the frequency of publication of such papers in the 2020s, while ViT and contrastive/few-shot dehazing has gained increased recognition from 2021 to 2023. The number of ViT-based dehazing surges in 2023 visually demonstrates that there is a shift in focus from using CNN to ViT for image dehazing owing to its novelty and promising results relative to CNN-based approaches. The same argument applies to contrastive/few-shot dehazing works. 

\begin{figure}[hbt!]
    \centering
    \includegraphics[scale=0.80]{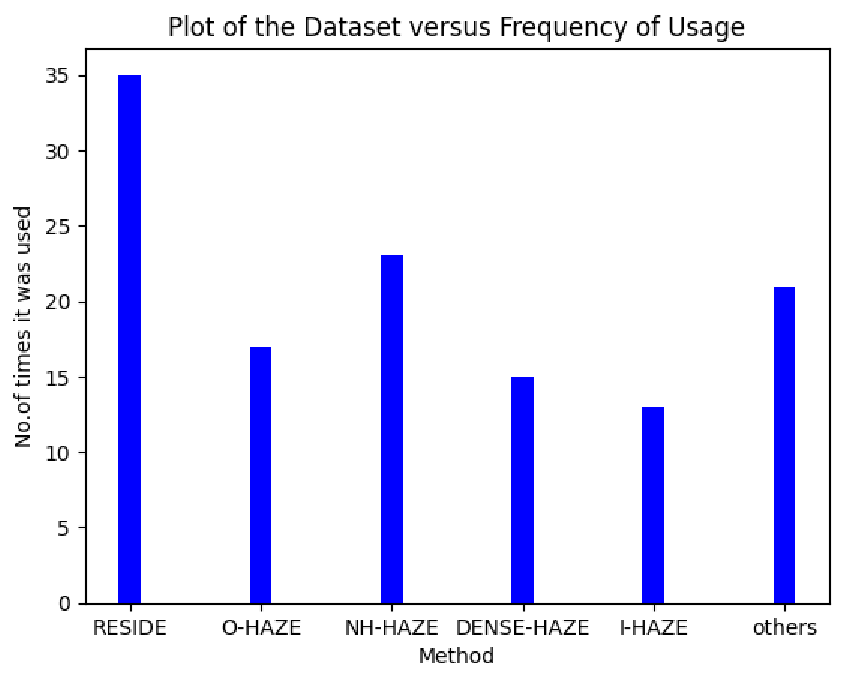}
    \caption{Bar graphs of number of times a proposed dehazing approach utilized the commonly dehazing benchmarked datasets (RESIDE, NH-HAZE, DENSE-HAZE, O-HAZE), along with the non-benchmark (`others') for all the works discussed so far. We can see that RESIDE remained the most widely utilized, followed by NH-HAZE, then `others', O-HAZE, DENSE-HAZE and lastly the I-HAZE.}
    \label{Bar_Plots2}
\end{figure}

Since 2024 is the current year of writing, the number of corresponding published works is obviously lower, but we foresee a continual increment in such papers to come. Based on the review thus far, a few discussion points are as follows:

\begin{itemize}
    \item Although there are few studies on utilizing a GCN using the CNN architecture approach, there is little to no work currently on combining GCN with the ViT architecture for dehazing. Apart from hinting at a possible and potential future research direction, utilizing graphs can provide either global or local information about the image. For local information, their interaction can be defined by an undirect graph adjacency matrix, and to incorporate such locality into a ViT, one way is to insert a spectral GCN for simple local feature extraction via matrix and multiplication, and combining it with the SE blocks to model the inter-dependencies between the channels in the feed-forward phase of the ViT \cite{jin2022gvit}. For global information, the dual GCN module described by Hu et.al. \cite{hu2023hazy} can be modified to suit ViT architecture.
    
    \item The RESIDE dataset remains the most popular benchmark dataset for evaluation, regardless of the proposed dehazing approach proposed. As mentioned in the introduction, it is one of the first large-scale image dehazing datasets, comprising 110,500 synthetic haze images and 4807 real hazy images, hence even ViT-based approaches can attain high performance metrics by using them. The O-HAZE, NH-HAZE, and DENSE-HAZE are also highly utilized benchmarks next to RESIDE. Figure \ref{Bar_Plots2} depicts the bar graphs of the frequency of usage of each type of dehazing benchmark based on our review so far, with `others' depicting datasets that are not aforementioned as above. From the figure we can quantitatively showed our aforementioned claim that RESIDE is the most utilized benchmark, followed by NH-HAZE, then `others', O-HAZE, DENSE-HAZE and lastly the I-HAZE.
    
    \item The increased in the number of datasets for evaluation that are categorized under `others' demonstrated that emerging works have begun to go beyond the aforementioned benchmarks and make the evaluations of their proposed algorithms more holistic. Some of the studies, such as those by Wei et.al. \cite{wei2022zero}, and ST-UNet by Bian et.al. \cite{bian2022swin} were applied to dehazing for remote sensing, illustrating the increased focus on the applicability of dehazing, rather than merely comparing the proposed algorithm to the others on the benchmark datasets.
    
\end{itemize}

\subsection{Performance evaluation on benchmark datasets: NH-HAZE, DENSE-HAZE, I-HAZE, and O-HAZE}

\begin{figure}[hbt!]
    \centering
    \includegraphics[scale=0.70]{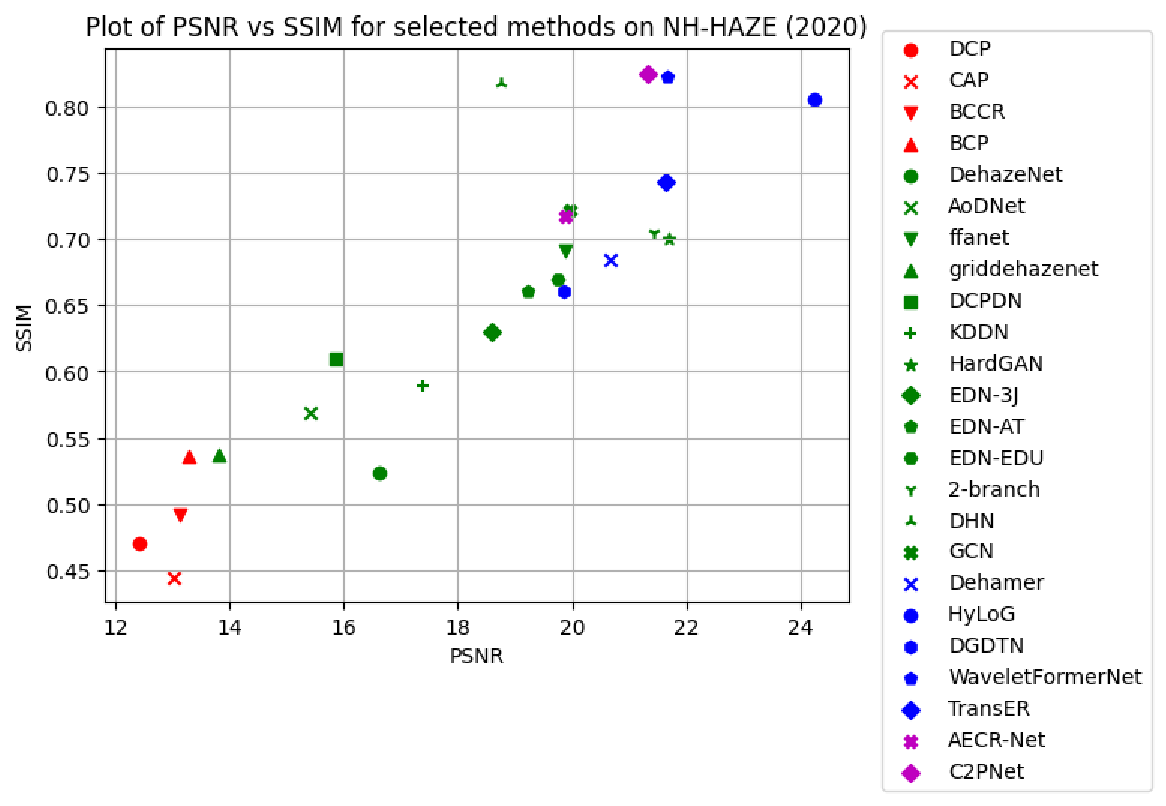}
    \caption{Scatter plots of the PSNR vs SSIM for the selected dehazing methods displayed in Table 7 as applied on NH-HAZE (2020). Plots in red denote prior-based dehazing, plots in green denote CNN-based dehazing, plots in blue denote ViT-based dehazing, and plots in magenta denote contrastive or few-shot-based dehazing.}
    \label{nh_haze_scatter}
\end{figure}

\begin{figure}[hbt!]
    \centering
    \includegraphics[scale=0.80]{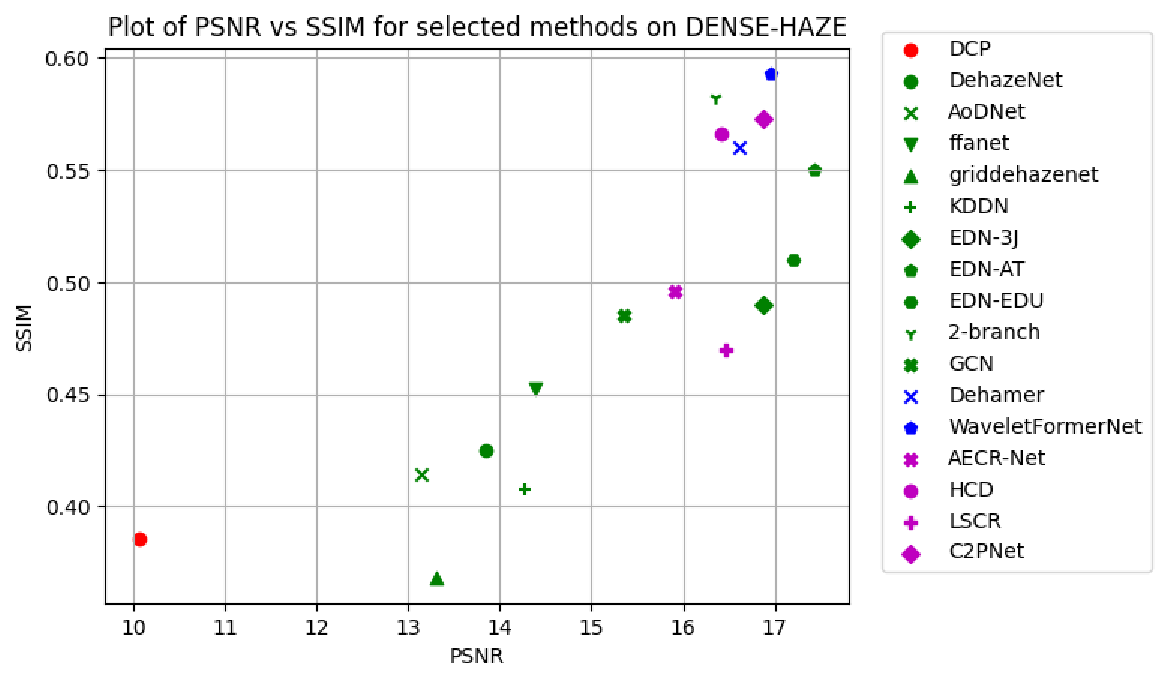}
    \caption{Scatter plots of the PSNR vs SSIM for the selected dehazing methods displayed in Table 8 as applied on DENSE-HAZE. Plots in red denote prior-based dehazing, plots in green denote CNN-based dehazing, plots in blue denote ViT-based dehazing, and plots in magenta denote contrastive or few-shot-based dehazing.}
    \label{dense_haze_scatter}
\end{figure}

\begin{figure}[hbt!]
    \centering
    \includegraphics[scale=0.80]{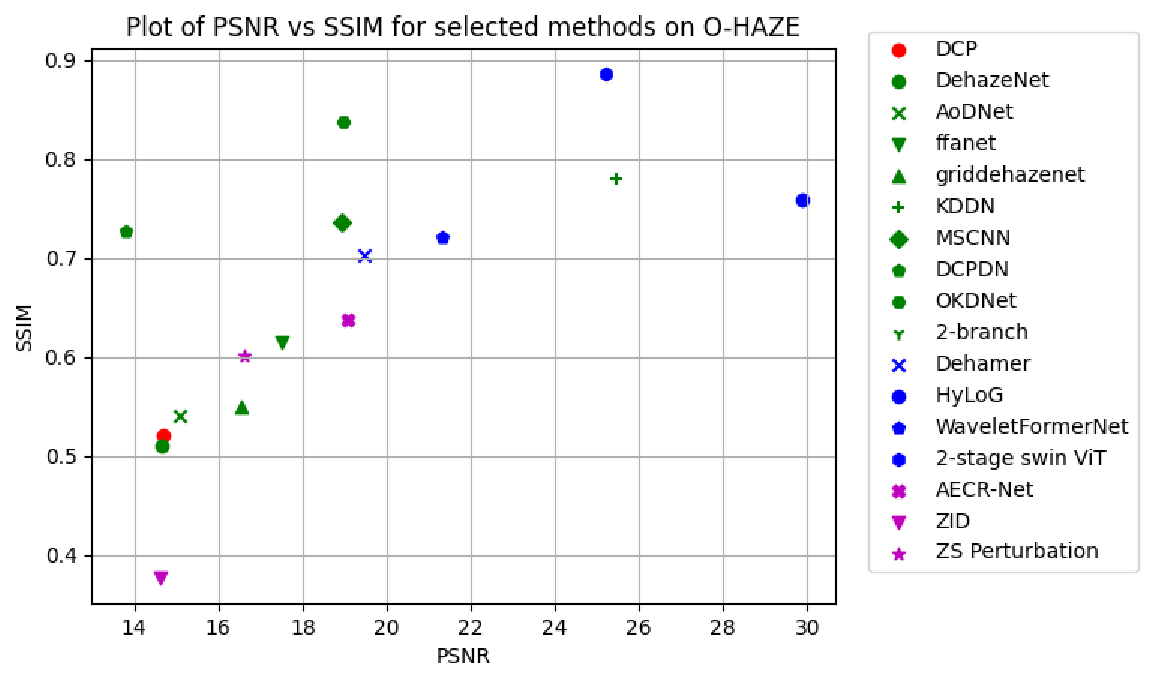}
    \caption{Scatter plots of the PSNR vs SSIM for the selected dehazing methods displayed in Table 9 as applied on O-HAZE. Plots in red denote prior-based dehazing, plots in green denote CNN-based dehazing, plots in blue denote ViT-based dehazing, and plots in magenta denote contrastive or few-shot-based dehazing.}
    \label{o_haze_scatter}
\end{figure}

\begin{figure}[hbt!]
    \centering
    \includegraphics[scale=0.80]{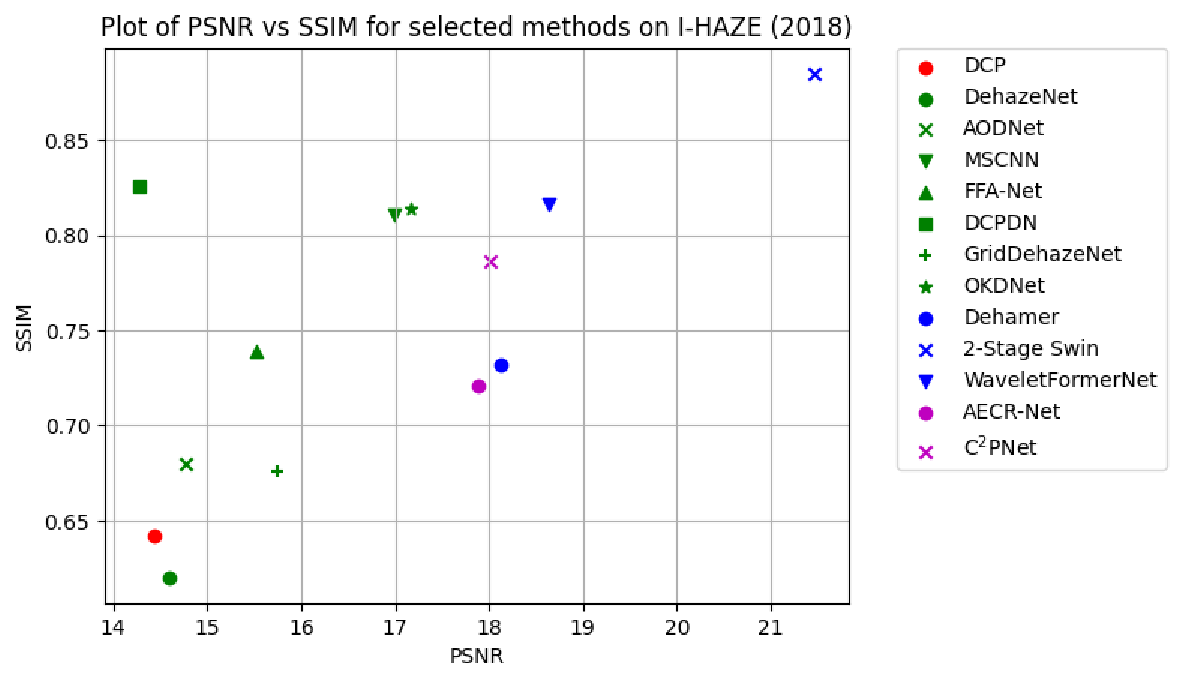}
    \caption{Scatter plots of the PSNR vs SSIM for the selected dehazing methods displayed in Table 10 as applied on I-HAZE. Plots in red denote prior-based dehazing, plots in green denote CNN-based dehazing, plots in blue denote ViT-based dehazing, and plots in magenta denote contrastive or few-shot-based dehazing.}
    \label{i_haze_scatter}
\end{figure}

Tables 7, 8, 9, and 10 depict the tabulated dehazing metrics for the selected methods on NH-HAZE (2020), DENSE-HAZE, O-HAZE and I-HAZE, respectively. A plot of the PSNR vs SSIM for the methods in the corresponding datasets is also depicted in Figure \ref{nh_haze_scatter}, \ref{dense_haze_scatter}, \ref{o_haze_scatter} and \ref{i_haze_scatter} respectively. We chose the NH-HAZE benchmark proposed in 2020 for the numerical comparison because it is the most widely used non-homogeneous haze dataset relative to the 2021 and 2023 version. 

The first trend we can observed is that for all the datasets described, the results from the ViT-based approaches are always at the higher-end of the respective PSNR vs SSIM plot, illustrating their promising and superior dehazing capabilities relative to CNN-based dehazing in general. The prior-based approaches do not fare well in general, compared to the rest of approaches, as best seen in the NH-HAZE scatter plots. The contrastive and few-shot approaches displayed mixed results, depending on the type of dataset utilized. For the NH-HAZE and DENSE-HAZE, some approaches, such as the C$^{2}$PNet consistently outperformed CNN-based dehazing in general, while others, such as AECR-Net, fared better in NH-HAZE than in DENSE-HAZE. The selected approaches did not perform well relative to the ViT-based approaches for O-HAZE. However, for I-HAZE, the contrastive and few-shot approaches performed better than CNN-based dehazing in terms of PSNR, but not SSIM.

The second trend we can obviously notice is that all the numerical results obtained from DENSE-HAZE are lower than those of O-HAZE, I-HAZE and NH-HAZE, regardless of the dehazing approach. DENSE-HAZE, as stated by the original authors of the dataset, was purposely designed to test the limits of SOTA dehazing architectures and to focus on the problem of thick haze removal. The highest PSNR and SSIM obtained for this dataset is the WaveletFormerNet, with PSNR and SSIM of 16.95 and 0.5930, respectively. To the best of our knowledge, there are no existing works  that push the PSNR and SSIM values to beyond 18.00 and 0.6500 respectively. Although the visual outputs in this case are not very pleasing, it can potentially pave the first step for realizing effective thick haze removal algorithms, which can be useful in remote sensing satellites and UAVs in analyzing scenery via the RGB spectrum under extreme haze scenarios from serious forest fires. This can complement the corresponding analysis from the infrared sensor data, which is more easily obtained as infrared radiation can easily transmit through thick haze because of its lower wavelength.

% NH-HAZE.

\begin{table*} 
\caption{Results of the selected dehazing models evaluated on the NH-HAZE (2020) Dataset using the PSNR and SSIM as metrics.}
\begin{tabular}{p{5cm}p{3cm}p{2cm}p{2cm}}
\hline
\textbf{Works} & \textbf{Type} & \textbf{PSNR} & \textbf{SSIM} \\
\hline 
DCP \cite{he2010single} & Prior & 12.40 & 0.4710 \\
CAP \cite{zhu2015fast} & Prior & 13.01 & 0.4450 \\
BCCR \cite{meng2013efficient} & Prior & 13.13 & 0.4920\\
BCP+DCP \cite{li2023single} & Prior  & 13.29 & 0.5360 \\ 
Density Class \cite{yang2023single} & Prior & 12.39 & 0.5921\\
\hline
DehazeNet \cite{cai2016dehazenet} & CNN & 16.62 & 0.5238 \\
AOD-Net \cite{li2017aod} & CNN & 15.40 & 0.5693 \\
FFA-Net \cite{qin2020ffa} & CNN & 19.87 & 0.6915 \\
GridDehazeNet \cite{liu2019griddehazenet} & CNN & 13.80 & 0.5370 \\
DCPDN \cite{zhang2018densely} & CNN & 15.86 & 0.6100 \\
KDDN \cite{hong2020distilling} & CNN & 17.39 & 0.5897\\
HardGAN \cite{deng2020hardgan} & CNN & 21.70 & 0.7000\\
EDN-3J \cite{yu2020ensemble} & CNN & 18.58 & 0.6300 \\
EDN-AT \cite{yu2020ensemble} & CNN & 19.22 & 0.6600 \\
EDN-EDU \cite{yu2020ensemble} & CNN & 19.76 & 0.6700 \\
2-branch \cite{yu2021two} & CNN & 21.44 & 0.7040 \\
DHN \cite{jiang2023deep} & CNN & 18.76 & 0.8180\\
GCN \cite{hu2023hazy} & GCN & 19.96 & 0.7216\\
\hline
Dehamer \cite{guo2022image} & ViT & 20.66 & 0.6844 \\
HyLoG \cite{zhao2021complementary} & ViT & 24.26 & 0.8050 \\
DGDTN \cite{zhang2023deep} & ViT & 19.86 & 0.6600 \\
WaveletFormerNet \cite{zhang2024waveletformernet} & ViT & 21.68 & 0.8220\\
TransER \cite{hoang2023transer} & ViT & 21.64 & 0.7430 \\
\hline
AECR-Net \cite{wu2021contrastive} & CoFS & 19.88 & 0.7170\\
C$^{2}$PNet \cite{zheng2023curricular} & CoFS & 21.32 & 0.8250 \\
\hline
\end{tabular}
\label{tab:table7}
\end{table*}

% DENSE-HAZE

\begin{table*} 
\caption{Results of the selected dehazing models evaluated on the DENSE-HAZE Dataset using the PSNR and SSIM as metrics.}
\begin{tabular}{p{5cm}p{3cm}p{2cm}p{2cm}}
\hline
\textbf{Works} & \textbf{Type} & \textbf{PSNR} & \textbf{SSIM} \\
\hline
DCP \cite{he2010single} & Prior & 10.06 & 0.3856 \\
\hline 
DehazeNet \cite{cai2016dehazenet} & CNN & 13.84 & 0.4252 \\
AOD-Net \cite{li2017aod} & CNN & 13.14 & 0.4144 \\
FFA-Net \cite{qin2020ffa} & CNN & 14.39 & 0.4524 \\
GridDehazeNet \cite{liu2019griddehazenet} & CNN & 13.31 & 0.3681 \\
KDDN \cite{hong2020distilling} & CNN & 14.28 & 0.4074\\
EDN-3J \cite{yu2020ensemble} & CNN & 16.87 & 0.4900 \\
EDN-AT \cite{yu2020ensemble} & CNN & 17.44 & 0.5500\\
EDN-EDU \cite{yu2020ensemble} & CNN & 17.21 & 0.5100 \\
2-branch \cite{yu2021two} & CNN & 16.36 & 0.5820 \\
GCN \cite{hu2023hazy} & GCN & 15.36 & 0.4853\\
\hline
Dehamer \cite{guo2022image} & ViT & 16.62 & 0.5602\\
WaveletFormerNet \cite{zhang2024waveletformernet} & ViT & 16.95 & 0.5930 \\
\hline
AECR-Net \cite{wu2021contrastive} & CoFS & 15.91 & 0.4960\\
HCD \cite{wang2024restoring} & CoFS & 16.41 & 0.5662 \\
LSCR \cite{bai2022contrastive} & CoFS & 16.47 & 0.4700 \\
C$^{2}$PNet \cite{zheng2023curricular} & CoFS & 16.88 & 0.5730\\
\hline
\end{tabular}
\label{tab:table8}
\end{table*}

% O-HAZE.

\begin{table*} 
\caption{Results of the selected dehazing models evaluated on the O-HAZE Dataset using the PSNR and SSIM as metrics.}
\begin{tabular}{p{5cm}p{3cm}p{2cm}p{2cm}}
\hline
\textbf{Works} & \textbf{Type} & \textbf{PSNR} & \textbf{SSIM} \\
\hline
DCP \cite{he2010single} & Prior & 14.68 & 0.5200 \\
\hline 
DehazeNet \cite{cai2016dehazenet} & CNN & 14.65 & 0.5100 \\
AOD-Net \cite{li2017aod} & CNN & 15.07 & 0.5400 \\
MSCNN \cite{ren2020single} & CNN & 18.94 & 0.7359 \\
FFA-Net \cite{qin2020ffa} & CNN & 17.52 & 0.6140 \\
GridDehazeNet \cite{liu2019griddehazenet} & CNN & 16.53 & 0.5500 \\
DCPDN \cite{zhang2018densely} & CNN & 13.79 & 0.7260 \\
KDDN \cite{hong2020distilling} & CNN & 25.46 & 0.7800 \\
OKDNet \cite{lan2022online} & CNN & 18.96 & 0.8370\\
2-branch \cite{yu2021two} & CNN & 25.20 & 0.8858 \\
\hline
Dehamer \cite{guo2022image} & ViT & 19.47 & 0.7020\\
HyLoG \cite{zhao2021complementary} & ViT & 29.87 & 0.7580\\
2-stage swin ViT \cite{li2022two} & ViT & 25.20 & 0.8858 \\
WaveletFormerNet \cite{zhang2024waveletformernet} & ViT & 21.32 & 0.7200\\
\hline
AECR-Net \cite{wu2021contrastive} & CoFS & 19.06 & 0.6370\\
ZID \cite{li2020zero} & CoFS & 14.60 & 0.3770 \\ 
ZS Perturbation \cite{kar2021zero} & CoFS & 16.63 & 0.6010\\
\hline
\end{tabular}
\label{tab:table9}
\end{table*}

% I-HAZE

\begin{table*} 
\caption{Results of the selected dehazing models evaluated on the I-HAZE Dataset using the PSNR and SSIM as metrics.}
\begin{tabular}{p{5cm}p{3cm}p{2cm}p{2cm}}
\hline
\textbf{Works} & \textbf{Type} & \textbf{PSNR} & \textbf{SSIM} \\
\hline
DCP \cite{he2010single} & Prior & 14.43 & 0.6420  \\
\hline 
DehazeNet \cite{cai2016dehazenet} & CNN & 14.59 & 0.6200 \\
AOD-Net \cite{li2017aod} & CNN & 14.77 & 0.6800 \\
MSCNN \cite{ren2020single} & CNN & 16.99 & 0.8109 \\
FFA-Net \cite{qin2020ffa} & CNN & 15.52 & 0.7390 \\
DCPDN \cite{zhang2018densely} & CNN & 14.27 & 0.8260\\
GridDehazeNet \cite{liu2019griddehazenet} & CNN & 15.75 & 0.6760 \\
OKDNet \cite{lan2022online} & CNN & 17.16 & 0.8140\\ 
\hline
Dehamer \cite{guo2022image} & ViT & 18.13 & 0.7320 \\
2-stage swin ViT \cite{li2022two} & ViT & 21.47 & 0.8849 \\
WaveletFormerNet \cite{zhang2024waveletformernet} & ViT & 18.64 & 0.8160 \\
\hline
AECR-Net \cite{wu2021contrastive} & CoFS & 17.89 & 0.7210 \\
C$^{2}$PNet \cite{zheng2023curricular} & CoFS & 18.02 & 0.7860 \\
\hline
\end{tabular}
\label{tab:table10}
\end{table*}

\subsection{Performance evaluation on RESIDE benchmark dataset}

\begin{figure}[hbt!]
    \centering
    \includegraphics[scale=0.80]{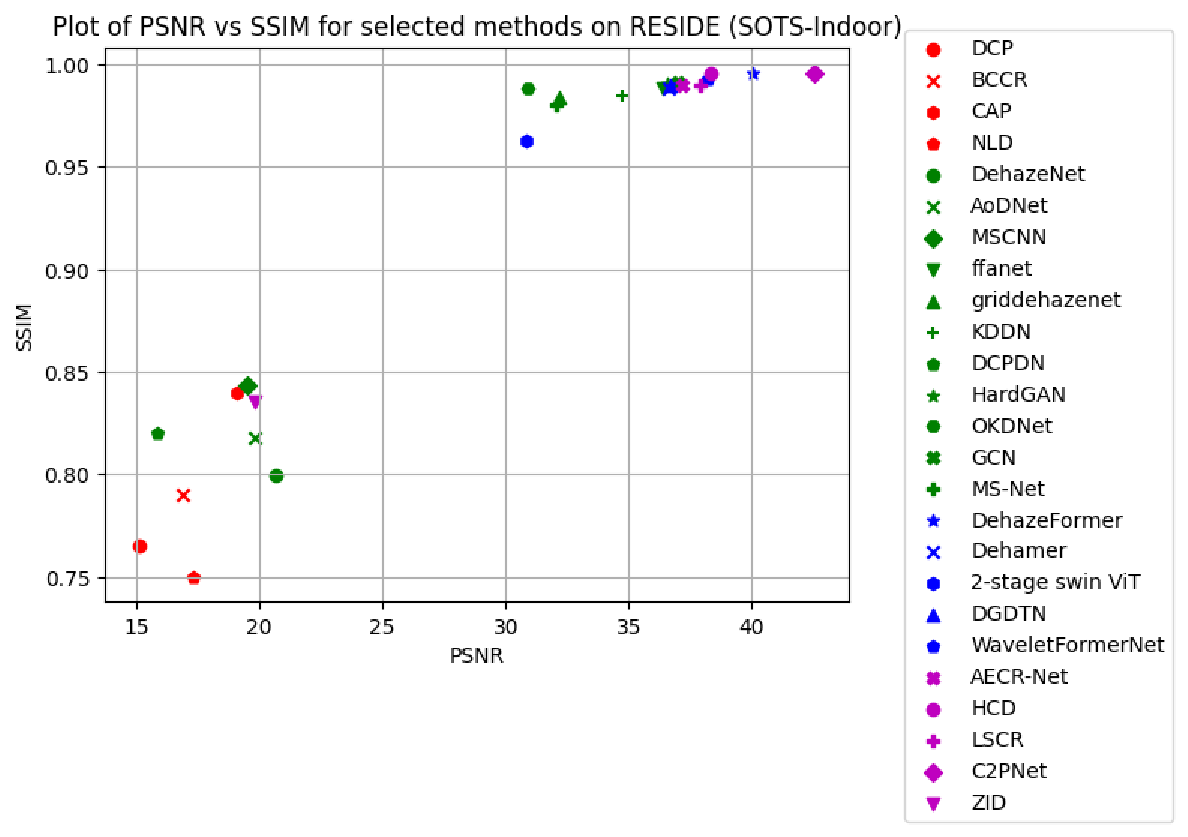}
    \caption{Scatter plots of the PSNR vs SSIM for the selected dehazing methods displayed in Table 11 as applied on RESIDE SOTS-Indoor. Plots in red denote prior-based dehazing, plots in green denote CNN-based dehazing, plots in blue denote ViT-based dehazing, and plots in magenta denote contrastive or few-shot-based dehazing.}
    \label{reside_indoor_haze_scatter}
\end{figure}

\begin{figure}[hbt!]
    \centering
    \includegraphics[scale=0.80]{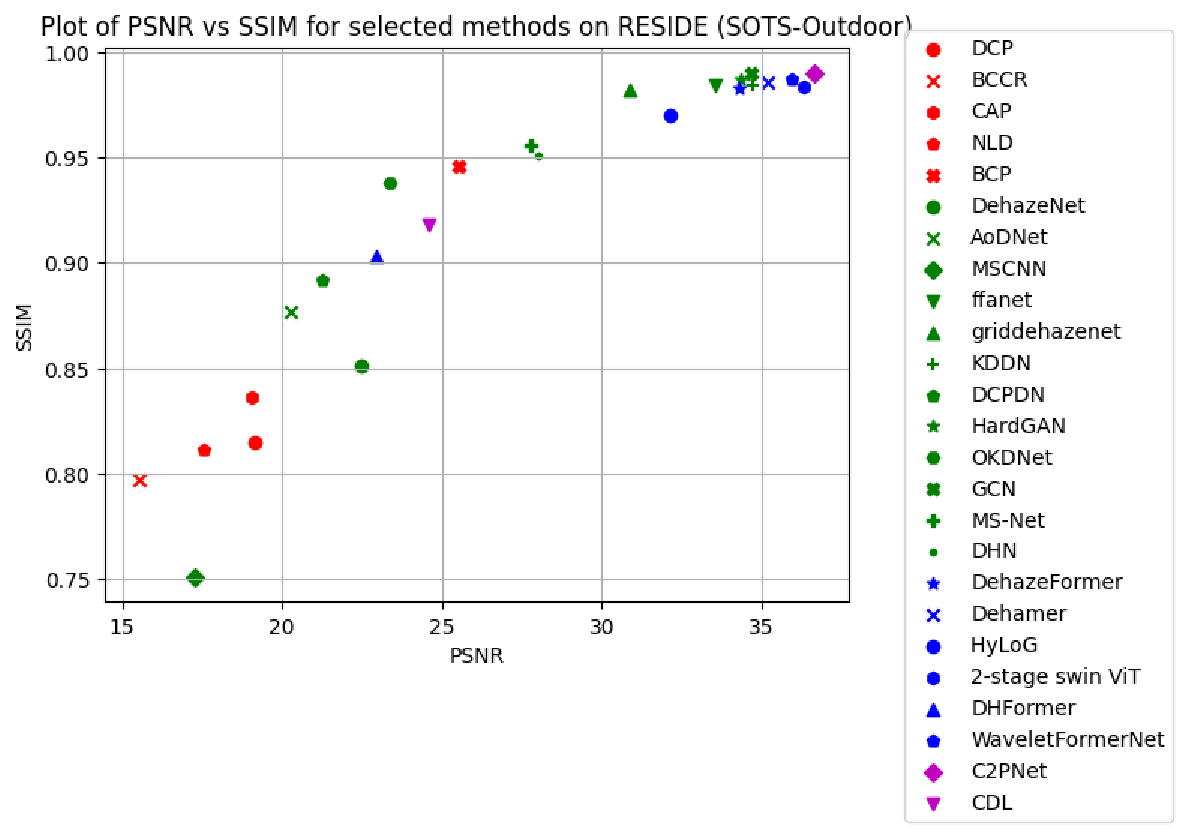}
    \caption{Scatter plots of the PSNR vs SSIM for the selected dehazing methods displayed in Table 12 as applied on RESIDE SOTS-Outdoor. Plots in red denote prior-based dehazing, plots in green denote CNN-based dehazing, plots in blue denote ViT-based dehazing, and plots in magenta denote contrastive or few-shot-based dehazing.}
    \label{reside_outdoor_haze_scatter}
\end{figure}

The RESIDE dataset is a bit more complex in that the evaluation or testing image subset can be categorized into three subsets, namely the Synthetic Objective Training Set (SOTS), which can be indoor (SOTS-Indoor) or outdoor (SOTS-Outdoor) images, or the Hybrid Subjective Testing Set HSTS. Table 11 and Figure \ref{reside_indoor_haze_scatter} depict the table of numerical PSNR and SSIM values for the selected approaches and their corresponding PSNR vs SSIM scatter plots for SOTS-Indoor respectively, while Table 12 and Figure \ref{reside_outdoor_haze_scatter} depict the corresponding table and plots for SOTS-Outdoor. Apart from observing that far better metric values were reported for a majority of the approaches relative to the previous datasets for both RESIDE subsets, CNN-based dehazing in this scenario competed with ViT-based dehazing and contrastive/few-shot dehazing to achieve optimal dehazing performance. Furthermore, even prior-based approaches can yield high metric values, with BCP attaining a new record of PSNR = 25.54 and SSIM = 0.9460 on the SOTS-Outdoor, on par with many of the learning-based approaches. This may be attributed to the fact that the haze in RESIDE is milder in intensity and more homogeneously distributed than that of the other datasets, rendering dehazing much simpler and more effective even for typical CNN approaches. However, there is a notable outlier in contrastive/few-shot dehazing in that the ZID does not perform exceptionally well relative to the rest of the dehazing approaches, as can be clearly seen in Figure \ref{reside_indoor_haze_scatter}. By comparing Figure \ref{reside_indoor_haze_scatter} and \ref{reside_outdoor_haze_scatter}, we can see that there is a large gap in the scatter plot of metrics values from SOTS-Indoor somewhere in the middle, unlike SOTS-Outdoor. This is because the metric values are more spread out in the SOTS-Outdoor dataset, while more of such values actually congregate in the upper-right corner of the plot in the SOTS-Indoor dataset, implying that many existing methods have the potential to excel in the indoor haze scenario. This is an interesting observation because typical indoor haze imagery does not invoked the airlight from the sun in the daytime, unlike outdoor haze imagery; hence, invoking the ASM may not yield effective dehazing performance in theory. However, as we have illustrated in the `Dehazing Approaches' section, a majority of proposed methods does not utilized the ASM and instead approach dehazing in an end-to-end manner, and therefore this also helps explain why intense research focuses have been invested to designing end-to-end dehazing architecture in recent times.

\begin{figure}[hbt!]
    \centering
    \includegraphics[scale=0.80]{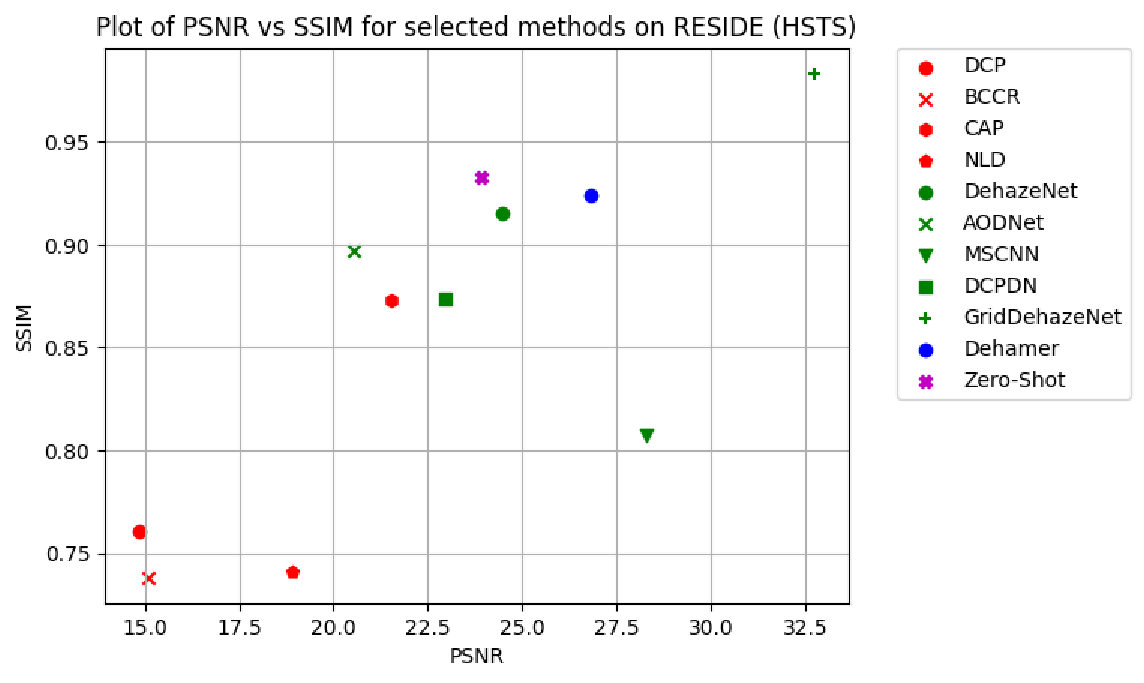}
    \caption{Scatter plots of the PSNR vs SSIM for the selected dehazing methods displayed in Table 13 as applied on RESIDE HSTS. Plots in red denote prior-based dehazing, plots in green denote CNN-based dehazing, plots in blue denote ViT-based dehazing, and plots in magenta denote contrastive or few-shot-based dehazing.}
    \label{reside_hsts_haze_scatter}
\end{figure}

Table 13 and Figure \ref{reside_hsts_haze_scatter} show the tabulated values and scatter plots for HSTS. As there are relatively fewer methods utilized for evaluation relative to the other RESIDE subsets, we cannot generalize in this case on which dehazing approaches are the best utilize, but we can once again observe that Dehamer, which is the sole ViT-based approach here, yielded high performance metrics relative to all prior-based works, and to a majority of CNN-based works, except for GridDehazeNet and MSCNN in terms of PSNR. \cite{wei2022zero}, which is the sole contrastive/few-shot method highlighted in this subset, has a higher SSIM value, but a lower PSNR value than that of the Dehamer. 

\begin{table*} 
\caption{Results of selected learning-based dehazing models evaluated with the RESIDE (SOTS-Indoor) Dataset using the PSNR and SSIM as metrics.}
\begin{tabular}{p{5cm}p{3cm}p{2cm}p{2cm}}
\hline
\textbf{Works} & \textbf{Type} & \textbf{PSNR} & \textbf{SSIM} \\
\hline 
DCP \cite{he2010single} & Prior & 15.09 & 0.7649 \\
BCCR \cite{meng2013efficient} & Prior & 16.88 & 0.7900 \\
CAP \cite{zhu2015fast} & Prior & 19.05 & 0.8400 \\
NLD \cite{berman2016non} & Prior & 17.29 & 0.7500 \\
\hline
DehazeNet \cite{cai2016dehazenet} & CNN & 20.64 & 0.7995\\
AOD-Net \cite{li2017aod} & CNN & 19.82 & 0.8178\\
MSCNN \cite{ren2020single} & CNN & 19.49 & 0.8436 \\
GridDehazeNet \cite{liu2019griddehazenet} & CNN & 32.16 & 0.9836\\
FFA-Net \cite{qin2020ffa} & CNN & 36.39 & 0.9886\\
DCPDN \cite{zhang2018densely} & CNN & 15.86 & 0.8200\\
KDDN \cite{hong2020distilling} & CNN & 34.72 & 0.9845\\
HardGAN \cite{deng2020hardgan} & CNN & 36.56 & 0.9905\\
MSNet \cite{yi2021msnet} & CNN & 32.04 & 0.9804 \\
OKDNet \cite{lan2022online} & CNN & 30.92 & 0.9880 \\
GCN \cite{hu2023hazy} & GCN & 37.01 & 0.9912 \\
\hline
DehazeFormer \cite{song2023vision} & ViT & 40.05 & 0.9960 \\
Dehamer \cite{guo2022image} & ViT & 36.63 & 0.9881 \\
DGDTN \cite{zhang2023deep} & ViT & 36.68 & 0.9900 \\
2-stage Swin ViT \cite{li2022two} & ViT & 30.84 & 0.9628\\ % RMB TO REPLOT
WaveletFormerNet \cite{zhang2024waveletformernet} & ViT & 38.20 & 0.9930 \\
\hline
AECR-Net \cite{wu2021contrastive} & CoFS & 37.17 & 0.9901\\
HCD \cite{wang2024restoring} & CoFS & 38.31 & 0.9954\\
LSCR \cite{bai2022contrastive} & CoFS & 37.93 & 0.9900 \\
C$^{2}$PNet \cite{zheng2023curricular} & CoFS & 42.56 & 0.9954\\
ZID \cite{li2020zero} & CoFS & 19.83 & 0.8353\\
\hline
\end{tabular}
\label{tab:table11}
\end{table*}

\begin{table*} 
\caption{Results of selected learning-based dehazing models evaluated with the RESIDE (SOTS-Outdoor) Dataset using the PSNR and SSIM as metrics.}
\begin{tabular}{p{5cm}p{3cm}p{2cm}p{2cm}}
\hline
\textbf{Works} & \textbf{Type} & \textbf{PSNR} & \textbf{SSIM} \\
\hline 
DCP \cite{he2010single} & Prior & 19.13 & 0.8148 \\
BCCR \cite{meng2013efficient} & Prior & 15.51 & 0.7970\\
CAP \cite{zhu2015fast} & Prior & 19.05 & 0.8360 \\
NLD \cite{berman2016non} & Prior & 17.57 & 0.8110 \\
BCP \cite{li2023single} & Prior & 25.54 & 0.9460 \\
\hline
DehazeNet \cite{cai2016dehazenet} & CNN & 22.46 & 0.8514\\
AOD-Net \cite{li2017aod} & CNN & 20.29 & 0.8765\\
MSCNN \cite{ren2020single} & CNN & 17.27 & 0.7510 \\
GridDehazeNet \cite{liu2019griddehazenet} & CNN & 30.86 & 0.9819\\
FFA-Net \cite{qin2020ffa} & CNN & 33.57 & 0.9840\\
DCPDN \cite{zhang2018densely} & CNN & 21.24 & 0.8920\\
MSNet \cite{yi2021msnet} & CNN & 27.78 & 0.9556 \\
KDDN \cite{hong2020distilling} & CNN & 34.72 & 0.9845\\
HardGAN \cite{deng2020hardgan} & CNN & 34.34 & 0.9871\\
OKDNet \cite{lan2022online} & CNN & 23.38 & 0.9380 \\
DHN \cite{jiang2023deep} & CNN & 28.01 & 0.9510\\
GCN \cite{hu2023hazy} & GCN & 34.69 & 0.9903 \\
\hline
DehazeFormer \cite{song2023vision} & ViT & 34.29 & 0.9830 \\
DHFormer \cite{wasi2023dhformer} & ViT & 22.93 & 0.9030 \\
Dehamer \cite{guo2022image} & ViT & 35.18 & 0.9860 \\
HyLoG \cite{zhao2021complementary} & ViT & 32.17 & 0.9700 \\
%DGDTN \cite{zhang2023deep} & ViT & 36.68 & 0.9900 \\
2-stage Swin ViT \cite{li2022two} & ViT & 36.34 & 0.9836\\  % RMB TO REPLOT
WaveletFormerNet \cite{zhang2024waveletformernet} & ViT & 35.96 & 0.9870 \\
\hline
%AECR-Net \cite{wu2021contrastive} & CoFS & 37.17 & 0.9901\\
CDL \cite{chen2022unpaired} & CoFS & 24.61 & 0.9180 \\
C$^{2}$PNet \cite{zheng2023curricular} & CoFS & 36.68 & 0.9900\\
\hline
\end{tabular}
\label{tab:table12}
\end{table*}

\begin{table*} 
\caption{Results of selected learning-based dehazing models evaluated with the RESIDE (HSTS) Dataset using the PSNR and SSIM as metrics.}
\begin{tabular}{p{5cm}p{3cm}p{2cm}p{2cm}}
\hline
\textbf{Works} & \textbf{Type} & \textbf{PSNR} & \textbf{SSIM} \\
\hline 
DCP \cite{he2010single} & Prior & 14.84 & 0.7609 \\
CAP \cite{zhu2015fast} & Prior & 21.53 & 0.8726 \\
BCCR \cite{meng2013efficient} & Prior & 15.08 & 0.7382\\
NLD \cite{berman2016non} & Prior & 18.92 & 0.7410 \\
\hline
DehazeNet \cite{cai2016dehazenet} & CNN & 24.48 & 0.9153 \\
AOD-Net \cite{li2017aod} & CNN & 20.55 & 0.8973 \\
MSCNN \cite{ren2020single} & CNN & 28.29 & 0.8070 \\
GridDehazeNet \cite{liu2019griddehazenet} & CNN & 32.75 & 0.9830\\
DCPDN \cite{zhang2018densely} & CNN & 22.94 & 0.8740\\
\hline
DHFormer \cite{wasi2023dhformer} & ViT & 26.83 & 0.9240 \\
\hline
Wei et.al. \cite{wei2022zero} & CoFS & 23.92 & 0.9330 \\
\hline
\end{tabular}
\label{tab:table13}
\end{table*}

The comparative illustration of the dehazed outputs from the selected approaches on the benchmarked datasets is covered in the appendix section.

\section{Image dehazing techniques for remote sensing applications}

In this section, we review remote sensing-based dehazing approaches, mainly classifying them into four categories: hyperspectral-based dehazing, very high-resolution (VHR)-based dehazing, Synthetic Aperture Radar (SAR)-based dehazing, and UAV imagery-based dehazing.

\subsection{Dehazing methods for hyperspectral imagery}

Using satellites, it is possible to capture and analyze remote sensing imagery and signals in other portions of the electromagnetic spectrum, such as the ultraviolet (UV) and infrared (IR) domains. Specifically, the infrared spectrum can be divided into three regions: near- (NIR), mid- (MIR), and far-infrared (FIR). Images captured in these ranges are referred to as Hyperspectral Images (HSI), and contain more spectral information than the other two platforms. Because they are of essential importance in environmental and earth-science-based research and monitoring, image dehazing can complement infrared sensors in cross-checking possible changes and anomalies at the surface in presence of thick clouds or smoke (as in forest fires).

Makarau et.al. \cite{makarau2014haze} presented an in-homogeneous haze detection approach as well as a refined dark-object subtraction for Haze Transmission Map (HTM) computation, which permits a spectrally consistent multispectral haze removal procedure on both calibrated and uncalibrated satellite image data. Some of the hyperspectral datasets that were evaluated included Landsat 8 OLI and WorldView-2. Ma et.al. \cite{ma2022spectral} introduced a Spectral Grouping Network (SG-Net) that optimized the utilization of each spectral band during the dehazing reconstruction process. The network first categorizes each HSI into several band subsets using their intra-spectral correlations, followed by their convolution to extract the essential features in a parallel manner using multiple branches. An attention block also linked the adjacent network branches which selected useful information (i.e., information contained in longer bands not deteriorated by haze) to enable better feature transmission from each subset and hence enhance the reconstruction of the dehazed HSI. The Alternating Direction Method of Multipliers (ADMM)-Adam theory \cite{lin2021admm}, originally utilized for other HSI inverse problems such as image inpainting, was modified to solve the dehazing problem by Tang and Lin \cite{tanglinadmm2022}. The benefits of such an approach are that effective performances can be obtained just by using a small dataset. A U-Net architecture was implemented to obtain the usual deep learning-based output, which was complemented by a convex optimization approach as part of the (ADMM)-Adam theory. The same authors also proposed a transformer-driven IPT-based Hyperspectral dehazing (T2HyDHZ) paradigm \cite{tang2024transformer} that reformulated the hyperspectral dehazing problem as a spectral super-resolution problem. This is done by selecting the uncorrupted band carrying the most informative features, for which the super-resolution algorithm will spectrally upsample these bands in the feature space to arrive at a clean HSI output. A deep transformer network that captures non-local information in the spatial domain is then implemented to refine the HSI intermediate outputs. As of writing, this is the first spatial–spectral transformer network that can be applied to the HSI dehazing domain.

A hyperspectral-guided image dehazing GAN algorithm (HIDeGan) was proposed by Mehta et.al. \cite{mehta2020hidegan} which used an enhanced cycleGAN architecture (R2HCYCLE) and enhanced conditional GAN model (H2RGAN). By examining the complete spectrum, R2HCYCLE utilized HSI in conjunction with cycle consistency and skeletal loss to increase the quality of information recovery. Based on the HSI output produced by the R2HCYCLE, the H2RGAN then computes and restores the haze-free RGB image. An Asymmetric Attention Convolution Network (AACNet) was proposed by Xu et.al. \cite{xu2023aacnet} comprising of numerous Residual Asymmetric Attention Groups (RAAGs) with Residual Asymmetric Attention Blocks (RAABs) as key modules. With this architecture, low-frequency regions are skipped and the hazy regions are more prominently displayed, allowing for the maximum usage of deep spatial-spectral information provided by the network. A Pooling Channel Self-Attention (PCSA) allows smooth rebuilding of the spectral response curve impacted by haze. A contrastive haze-aware learning-based dynamic approach has been proposed by Nie et.al. \cite{nie2022contrastive} which utilized a contrastive clustering approach to learn the haze representation in an unsupervised manner, thereby allowing for the identification and discrimination of the unique haze state in each image. The acquired haze representation is complemented by a parameter generator that generates haze-aware parameters for the dynamic construction of the dehazing model for each hazy image, thereby enabling adaptive dehazing and enhancing its generalization capacity.

\subsection{Dehazing methods for VHR imagery}

Satellite sensors specifically engineered to capture images with an exceptionally high degree of detail are commonly used to acquire VHR imagery. The advancement of optical sensor technology has resulted in increasingly better spatial resolution, which raises the bar for the amount of detail that can be captured in these photographs. It goes without saying that designing effective dehazing algorithms is essential for subsequent higher-level visual tasks such as scene classification, segmentation and object detection or tracking.

Li et.al. \cite{li2023rsid} introduced an efficient and lightweight dehazing approach that utilized attention with smooth dilated convolution, which was incorporated into the encoder structure of their network. Such a design helps to achieve imbalanced management of hazy images with varying opacity, while lowering the number of parameters. In addition, channel weight fusion self-attention is incorporated into the decoder component to enable automatic learning and computation of different receptive field features in a pixel-by-pixel manner. Their work was motivated by the fact that remote sensing images are usually large-scale and contain rich information, which can render their processing demanding in computational terms. Wei et.al. \cite{wei2023self} introduced a self-supervised image dehazing architecture that utilized a zero-shot learning approach. Their model consists of three phases. The first involved using the DCP to process the hazy image and obtain its transmission maps, ambient light values and the DCP dehazed images. Two CNN-based RefineNets are then introduced in the second phase to enhance the DCP outputs. In the last phase, the hazy images are generated via the refined maps and enhanced dehazed images via the ASM. A novel cycle-consistency loss function was also devised to facilitate the zero-shot dehazing training process, and their model was tested in both homogeneous and non-homogeneous remote sensing hazy settings. 

A Gradient-based Haze Removal Network (GHRN) was highlighted by Ma et.al. \cite{ma2023incorporating} for tackling auxiliary VHR image-based dehazing. When performing VHR haze removal, taking into account auxiliary information might yield supplemental information to enhance valuable input features and hence lower the degree of uncertainty in the outcomes, as shown by Ma et.al. \cite{ma2022deep}. However, essential information in the auxiliary images may be distorted during dehazing as they may not match those of the target haze images in terms of perspectives, alignment and changes in land cover. Therefore, the GHRN utilized gradient information to not only highlight land cover boundaries but also to align extracted features, which were obtained via hierarchical convolution of the gradient images between that of the target and the auxiliary using an hourglass structure. An Iterative dehazing method for single remote sensing image (IDeRs) was proposed by Xu et.al. \cite{xu2019iders} and is one of the first studies to highlight the differences between natural and remote sensing image dehazing. Specifically, in the latter, because the sensors attached to the satellites are further away from the Earth's surface, atmospheric conditions can heavily affect the image quality in a wider variety of ways. Even for the same object detected, if the latter is covered by possible translucent contamination, it can easily be mistaken for haze. The IDeRs realized that the transmission maps of both natural haze and remote sensing images can be expressed by the same mathematical expressions and are related ti the optical depth of the image. The authors modified this depth to include a novel `virtual depth', which measured the amount of the coverage of the earth's surface due to haze, and also provided the image foreground-background distinction. The IDeRs also performed the dehazing process in an iterative manner so that haze from a wide variety of scales could be removed step-by-step. Finally, to address artifacts left behind owing to halos and over-saturation, a fusion module that fuses pixels and patch-wise dehazing operators is also incorporated in the model. 

An end-to-end trainable attentive transformer network for dehazing satellite aerial images was proposed by Kulkarni and Murala \cite{kulkarni2023aerial}. Their motivation was that previous existing techniques either required extra prior information during training, or yielded sub-optimal results for varying haze as the derived features lacked both local and global interdependence. While the transformer network handles global inter-dependency, their deformable convolution, which employs a spatially attentive offset extractor, concentrates on pertinent contextual information, which handles local inter-dependency. To enable the easier transfer of edge features from the shallower to deeper layers of the network, edge-boosting skip connections were incorporated. 

\subsection{Dehazing methods SAR imagery}

Radar-based remote sensing works by taking uses the emission and reception of electromagnetic waves on Earth. These methods are capable of capturing high-spatial-resolution images of scenery in any weather. Therefore, Synthetic Aperture Radar (SAR)-based images have been utilized in diverse fields, and it is no surprise that they have also been used in dehazing applications.

Huang et.al. \cite{huang2020single} introduced a novel fusion dehazing paradigm that includes an end-to-end conditional GAN (cGAN) and combined the information contained in the RGB and SAR images to minimize image blurs. The generator of the cGAN contained a dilated residual block module to enhance dehazing performance, and a new dataset (SateHaze1K) was created to address the lack of availability of hazy SAR-RGB images. Liu et.al. \cite{liu2024approach} proposed a combined optimization model of dehazing and SAR ship detection using a dehazing module that is self-adaptive and inserted before the lightweight detection architecture. In addition, the latter also invoked an improved YOLO v5s model \cite{jocher2020ultralytics}, as well as a deformable convolution module to improve the detection model’s performance.

\subsection{Dehazing methods for UAV imagery}

Zhang et.al. \cite{zhang2021uav} introduced a UAV image dehazing algorithm based on a frequency-domain saliency model (originally utilized to explain the visual attention mechanism \cite{hou2017deeply}) to highlight essential hazy regions for better transmission value estimation and correction, which is performed in a two-scale manner. A suppression parameter that suppresses any color distortions (in which the authors argued that such is common in UAV dehazing imagery) was also incorporated, and the resultant saliency map was computed by taking the weights of the transmission correction to reduce as much texture detail loss and color distortions as possible in the hazy highlights. The AEE-Net was proposed by Cai et.al. \cite{cai2021aee} which also emphasized the simplicity of the strcture and utilized the modified ASM, as first proposed by the AOD-Net. To collect the feature information of varying proportions of the hazy images, convolution kernels of various sizes were inserted into AEE-Net. One of the first dehazing GAN incorporating online distillation for model compression in UAV applications was introduced by Liu et.al. \cite{liu2022aerial}, under Aerial Image Dehazing Network Compression (AIDNC). A lightweight residual structure and channel pruning are used in the student network, and distillation is performed via a multi-granularity scheme. The network was optimized using guidance information from the gradually improved teacher network, which served as the generator.

An adaptive Siamese dehazing network was introduced by Sun et.al. \cite{sun2023adaptive} for video object tracking. To address the challenge of inadequate object tracking data in hazy scenarios, the authors collected both real-world and synthetic haze scenes to create four of their own datasets: the UAV123-AH, UAV20L-AH, UAVDT-AH, and DTB70-AH. The new adaptive module is based on the modified DCP, which preserves edge information in the dehazed images so that the object tracker can extract features with better characterization. A dynamic template update branch was also introduced to dynamically select more dependable template images during the tracking process, as the UAV images may be distorted and suffer from tracking drift due to shaking from external forces or interference from similar objects in the same scenery. It is interesting to note that earlier studies in UAV image dehazing mainly utilized the DCP and its modification, for instance Yu et.al. \cite{yu2019unmanned}, Liu et.al. \cite{liu2022new}, and Sha et.al. in \cite{sha2020unmanned}. For the latter, dehazing was used for better visual power inspection of the wind farm. 

Table 14 summarizes all the aforementioned remote sensing-based dehazing works, indicating the year in which the work was published, the domain in which the dehazing was performed, and what datasets were used.

\begin{table*} 
\caption{Summary of the aforementioned remote sensing-based dehazing works, indicating the year published, the domain in which the dehazing was done and what datasets were used.}
\begin{tabular}{p{4cm}p{1cm}p{2cm}p{6cm}}
\hline
\textbf{Works} & \textbf{Year} & \textbf{Domain} & \textbf{Some Datasets Used} \\
\hline
Makarau et.al. \cite{makarau2014haze} & 2014 & HSI & Landsat 8 OLI, WorldView-2\\
SG-Net \cite{ma2022spectral} & 2022 & HSI & Chikusei, Gaofen-5, Zhuhai-1\\
ADMM-Adam \cite{tanglinadmm2022} & 2022 & HSI & Airborne Visible/Infrared Imaging Spectrometer (AVIRIS)\\
T2HyDHZ \cite{tang2024transformer} & 2024 & HSI & AVIRIS\\
HIDeGan \cite{mehta2020hidegan} & 2020 & HSI & D-HAZY \cite{ancuti2016d}, RESIDE \cite{li2018benchmarking} \\
AACNet \cite{xu2023aacnet} & 2023 & HSI & EO-1 Hyperion, Landsat-8, Gaofen-5 \\
Nie et.al. \cite{nie2022contrastive} & 2022 & HSI & Sentinel-2, Gaofen-2 \\
\hline
Li et.al. \cite{li2023rsid} & 2023 & VHR & RESIDE \cite{li2018benchmarking}, Custom RS images \\
Wei et.al. \cite{wei2023self} & 2023 & VHR & SateHaze1k, D-LinkNet \cite{zhou2018d}, Custom RS images \\
GHRN \cite{ma2023incorporating} & 2023 & VHR & RRSSRD \cite{dong2021rrsgan}, WorldView2, Gaofen-2 \\
IDeRs \cite{xu2019iders} & 2023 & VHR & NWPU VHR-10 \cite{cheng2016survey}, NWPU-RESISC45 \cite{cheng2017remote}, RSOD \cite{long2017accurate}, IAILD \cite{maggiori2017can}, DOTA \cite{xia2018dota}\\
\hline
Huang et.al. \cite{huang2020single} & 2020 & SAR & SateHaze1K  \\
Liu et.al. \cite{liu2024approach} & 2024 & SAR & SeaShips, SeaShips-SMU   \\
\hline 
AEE-Net \cite{zhang2021uav} & 2021 & UAV & Outdoor Natural UAV images\\
Cai et.al. \cite{cai2021aee} & 2021 & UAV & Custom outdoor UAV images \\
AIDNC \cite{liu2022aerial} & 2022 & UAV & VisDrone \cite{zhu2018vision}, DOTA V2 \\
Sun et.al. \cite{sun2023adaptive} & 2023 & UAV & UAV123-AH, UAV20L-AH, UAVDT-AH,
DTB70-AH, UAVhaze \\
Yu et.al. \cite{yu2019unmanned} & 2019 & UAV & Custom outdoor UAV images \\
Liu et.al. \cite{liu2022new} & 2022 & UAV & Custom outdoor UAV images\\
Sha et.al. \cite{sha2020unmanned} & 2020 & UAV & Custom outdoor UAV images \\
\hline
\end{tabular}
\label{tab:table14}
\end{table*}

\section{Insights and trends in remote sensing and UAV-based dehazing}

We can gain some insights into the current trends in the field of remote sensing and UAV-based dehazing as follows:

\begin{itemize}
  \item For UAV-based dehazing works, numerous works have emphasized on lightweight but effective architectures (e.g., AEE-Net \cite{zhang2021uav}, AIDNC \cite{liu2022aerial}) as UAVs are resource and computationally limited. Model compression in the form of knowledge distillation is also a viable approach as demonstrated by AIDNC.
  \item Unlike other domains, there appear to be fewer studies on SAR-based dehazing. This issue can be due to the unavailability of large-scale SAR-based images in the public domain, since such dataset domains are typically confidential, as highlighted by \cite{song2022complex} in the few-shot classification domain. This makes the SAR-based dehazing task a major challenge, but also opens up the possibility of approaching the few-shot/contrastive paradigm in such areas, for which current works in this category are lacking.
  \item A myriad of dehazing approaches have been utilized, spanning all categories described in the previous sections. However, unlike the commonly benchmarked datasets, there are currently no GCN-based approaches for both satellite and UAV-based hazy images. As discussed in Hu et.al. \cite{hu2023hazy}, GCN is a good alternative to ViT in capturing global feature inter-dependency, which complements the CNN in computing local feature inter-dependency, and it would be interesting to compare the performances of such classes of methods to the ViTs in remote sensing, especially with works such as that of Kulkarni and Murala \cite{kulkarni2023aerial}. 
  
\end{itemize}

\section{Open problems and challenges}

Despite our extensive discussions, illustrations, and explanations of the existing SOTAs dehazing methods across the benchmarked dataset, as well as in the domain of remote sensing and UAV imagery, there are still concerns and challenges that need to be explored in further studies. The key issues to be highlighted are as follows:

\begin{itemize}
    \item Emphasis on efficiency
    \item Dehazing effectiveness to higher-level vision tasks
    \item Domain adaptation
    \item More emphasis on combining prior knowledge with deep learning
    \item More emphasis on detail recovery 
\end{itemize}

\subsection{Emphasis on efficiency}

ViT-based methods commonly outperformed CNN-based dehazing approaches in a myriad of benchmarked datasets. However, ViT is commonly more architecturally complex than CNN approaches, and hence usually requires higher training parameters, inference time, and computational resources. For example, DehazeFormer proposed five variants of its architecture: DehazeFormer-T, DehazeFormer-S, DehazeFormer-M, DehazeFormer-B, and DehazeFormer-L. The largest of the variants, DehazeFormer-L, required 25.44M parameters and 279.7G of Multiply Accumulate (MAC) operations. Therefore, for potential applications to mobile platforms such as UAVs, there is a need to design an efficient but effective dehazing architecture. As mentioned in the previous sections, one approach is to utilize model compression via knowledge distillation. Although not explicitly mentioned in the main sections, the Semi-UFormer proposed by Tong et.al. \cite{tong2022semi} is one of the recent work that combined distillation and the transformer, and can estimate the uncertainty of the pixel representations and leverage real-world haze information to guide the student network to produce the dehazed image more accurately. The same concepts could be extended to contrastive/few-shot learning-based dehazing, although to the best of our knowledge, there are few to no current works on such types. However, related work has been found in the area of image rain removal (image deraining), and one notable example is presented by Luo et.al. \cite{luo2023local} which combined a recurrent attention-distilling network and direction-enhanced contrastive learning to enhance the generalization capability of the model. Such a line of approaches could potentially be applied to the dehazing domain, as it involves an analogous local and global features interdependence computation along with a distillation paradigm.

\subsection{Dehazing effectiveness to higher-level vision tasks}

Ultimately, dehazing is a low-level image processing task as only basic haze features are extracted and computed for their removal. The wider purpose of dehazing is to provide higher-quality images for higher-level vision tasks such as image classification, scene understanding, object detection or tracking, and semantic or instance segmentation. Some of the studies described in remote sensing applications (e.g., Liu et.al. \cite{liu2024approach}) used dehazing for ship detection, and there are works that either utilized dehazing for segmentation (e.g., de Blois et.al. \cite{de2019learning}), or dehazing using complementary information from segmentation (e.g., Yu et.al. \cite{yu2022single}). However, it is still essential to note that there are emerging studies that cast doubt on the actual effectiveness of dehazing on CNN-based image classification, as discussed by Pei et.al. \cite{pei2018does}. According to them, a number of well-designed dehazing approaches may have little to no beneficial impact on the performance of the classification model, and occasionally would somewhat lower the classification accuracy. Therefore, more research could be dedicated to exploring the impact of dehazing on other vision applications, especially in mobile robotics platforms such as autonomous vehicles and UAVs. Some promising works in combining dehazing with Visual Odometry (VO) include Agarwal et.al. \cite{agarwal2014visual} (as applied with UAVs) and Li et.al. \cite{li2022high} (as applied to autonomous driving).

\subsection{Domain adaptation}

Despite the success of numerous dehazing approaches, it is important to note that some evaluations were performed on synthetic haze datasets. For instance, the RESIDE dataset comprises of 110,500 synthetic haze images and only 4807 real hazy images. The very high performance metrics as recorded in Figure \ref{reside_indoor_haze_scatter}-\ref{reside_hsts_haze_scatter}, as well as in Table 11-13, owe its success primarily to the ease of dehazing of the haze generated by the ASM, which is not reflective of how the real haze works. This can be justified by comparing the overall trends in Figure \ref{nh_haze_scatter}-\ref{i_haze_scatter} and Table 7-10 to that of the RESIDE tabulated values. An additional challenge arose when we realized that these real haze datasets do not contain as much data as RESIDE, hence partially explaining their lower performance metrics. 

This is the problem of domain adaptation, and the existing literature usually remedies this by making a more holistic comparisons of their approaches on a wide variety of datasets, as illustrated numerous times in this review. However, a question remains as to how to achieve an equivalent level of performance on RESIDE for other real-world haze datasets. One method is to introduce larger-scale real-haze datasets, as attempted by the multiple real fog images dataset (MRFID) \cite{liu2020end} and BeDDE \cite{zhao2020dehazing}. Another way is to design a more realistic haze model than the current ASM, but any additional complexities in the model manifested itself only in other lightning conditions (e.g., night-time haze), using the original ASM as a base. There may be a need for a more accurate physics-based dehazing approach, implying that prior-based approaches could be a viable option for further exploration. For instance, by utilizing the Density Class Prior as described in Table 3, a performance metric surpassing that of CNN-based dehazing was reported (PSNR = 12.73, SSIM = 0.9784), implying that utilizing priors may not inferior to that of learning-based approaches under careful designs and execution.

\subsection{More emphasis on combining prior knowledge with deep learning}

Although deep learning algorithms have demonstrated numerous successes in dehazing, several recent studies have shown that even prior statistical knowledge can complement deep learning-based network training. We have mentioned the success of well-designed priors in the previous subsection, but there are also emerging works on combining priors with CNN-based networks (e.g., Li et.al. \cite{li2021single}, Su et.al. \cite{su2021prior}, and Chen et.al. \cite{chen2023ipdnet}) and ViT-based network (e.g., VSPPA by Liu et.al. \cite{liu2023visual} and SwinTD-Net by Zhou et.al. \cite{zhou2023physical}). We believe that similar approaches could be applied to contrastive/few-shot learning approaches, and that such a paradigm could be applied more to satellite and UAV imagery. However, we would like to highlight that while designing good priors can assist the model to become less dependent on data and potentially increase its capacity for generalization, current research indicates that there is no consensus on which prior would be the best for a wide variety of haze scenarios. Therefore, combining priors with a myriad of learning networks would be a good research direction to push the current SOTA dehazing performances to a new level.

\subsection{More emphasis on detail recovery}

Most dehazing works have focused more on the haze removal step directly, rather than on the removal of haze artifacts that can occurred after the haze is removed. Because detail loss may be unavoidable during a standard dehazing operation, adding a distinct detail recovery network aids in restoring the intrinsic detail feature map, which can be subsequently fused with intermediate dehazed images to recover and improve the outputs. The observation that detail recovery is not focused upon is preliminary pointed out by Li et al. \cite{li2022single}, and this paradigm has only been applied in a few dehazing works as of current (e.g., the aforementioned work \cite{li2022single} and Fang et al. \cite{fang2022detail}). We have also pointed out that MSCNN is one of the few architectures that takes into account detail recovery via the fine-grained network, and it is a bit surprising that such a concept has been applied more in the related image deraining domain than in dehazing. (Some examples of deraining technique that utilized detail recovery include Gao et al. \cite{gao2022heavy}, Deng et al. \cite{deng2020detail}, Shen et al. \cite{shen2022detail} and Zhu et al. \cite{zhu2022hdrd}). Furthermore, to the best of our knowledge, there are few methods that concentrate on detail recovery for thick and non-homogeneous real haze scenarios (such as NH-HAZE and DENSE-HAZE). Considering that the techniques tested on the last two datasets performed worse than those on the previous homogeneous haze datasets, as seen repeatedly in this review, it would be interesting to investigate how the dehazing performance might be impacted in thick and heterogeneous haze when detail recovery is utilized.

\begin{figure}[hbt!]
    \centering
    \includegraphics[scale=0.55]{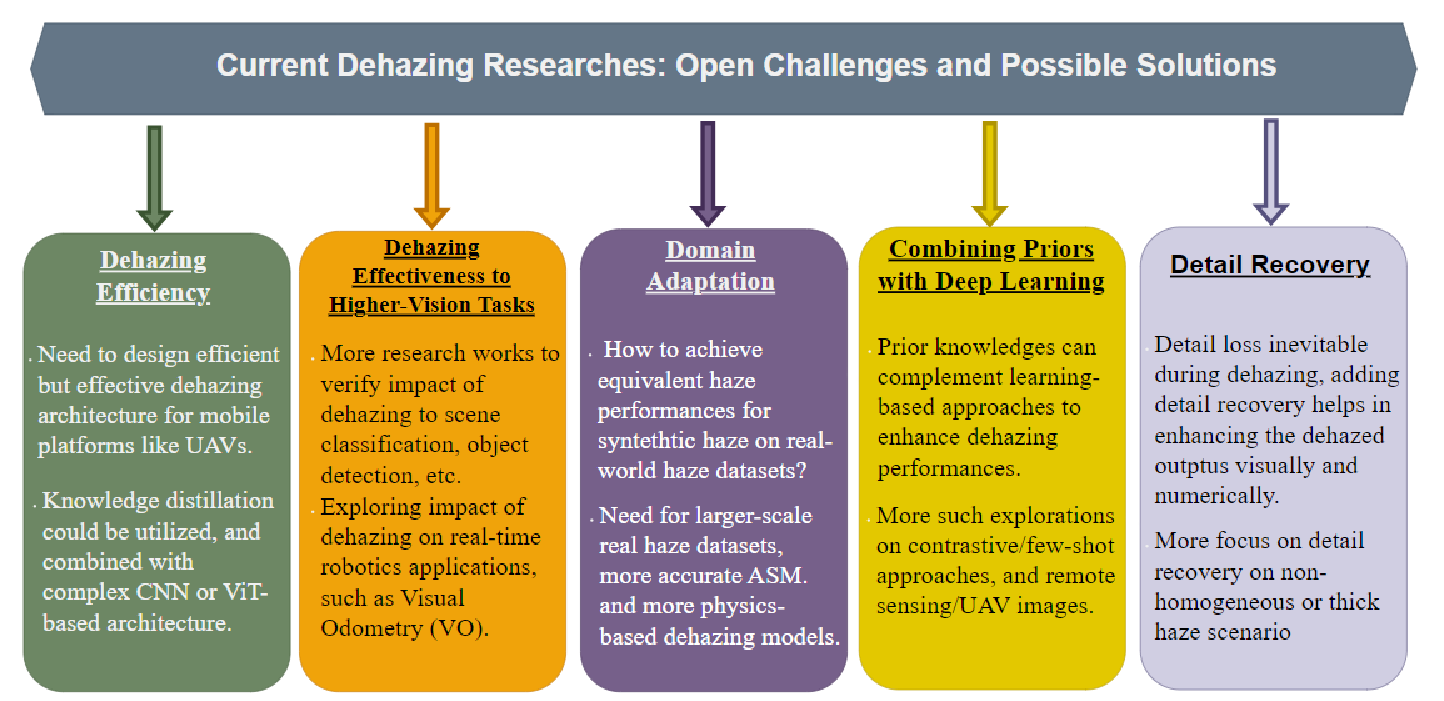}
    \caption{A schematic diagram depicting the summary of our discussions on the open challenges and possible solutions for current image dehazing research.}
    \label{Summary}
\end{figure}

Figure \ref{Summary} presents a summary of the open challenges and possible solutions discussed in this section via a schematic diagram, along with the essential discussions highlighted in a point-by-point format.

\section{Conclusions}
This review provides a comprehensive overview of the latest developments in image dehazing research, with a focus on remote sensing and UAV imagery. Haze can severely degrade image quality and interpretability in these fields, making dehazing a crucial step. Our review offers a valuable resource for researchers and engineers, featuring a taxonomy of dehazing approaches specific to various image types, including hyperspectral, high-resolution, SAR, and UAV imagery. We evaluated and compared the performance of different dehazing methods, including prior-based techniques, deep learning architectures, and emerging approaches like contrastive learning and few-shot learning. The review also identifies open problems, research gaps, and future directions, including the need for diverse real-world haze datasets, efficient algorithms, and integration of domain-specific knowledge into learning-based frameworks. This timely review provides a solid foundation for understanding the current state-of-the-art in image dehazing and charting future research directions. As the demand for high-quality images continues to grow, the development of accurate, efficient, and robust dehazing techniques will remain a critical research area with significant practical implications.

\section*{CRediT authorship contribution statement}

\textbf{Gao Yu Lee}: Conceptualization, Investigation, Methodology, Software, Writing - original draft, Writing- review $\&$ editing. \textbf{Jinkuan Chen}: Conceptualization, Investigation, Methodology, Software, Writing - original draft, Writing- review $\&$ editing. \textbf{Tanmoy Dam}:  Conceptualization, Investigation, Methodology, Supervision, Writing- review $\&$ editing. \textbf{Md Mefahul Ferdaus}: Supervision, Writing- review $\&$ editing. \textbf{Daniel Puiu Poenar}: Supervision. \textbf{Vu N. Duong}: Supervision.    

\section*{Declaration of Competing Interest}

The authors declare no competing interests or conflicts of interest that would influence the work reported in this review.

\section*{Acknowledgements}

This research/project is supported by the Civil Aviation Authority of Singapore and NTU under their collaboration in the Air Traffic Management Research Institute. Any opinions, findings and conclusions or recommendations expressed in this material are those of the author(s) and do not necessarily reflect the views of the Civil Aviation Authority of Singapore.

%\appendix

\section*{Appendix: Comparative Illustration of Dehazed Outputs}
\label{sec:sample:appendix}
\begin{figure}[hbt!]
    \centering
    \includegraphics[scale=0.50]{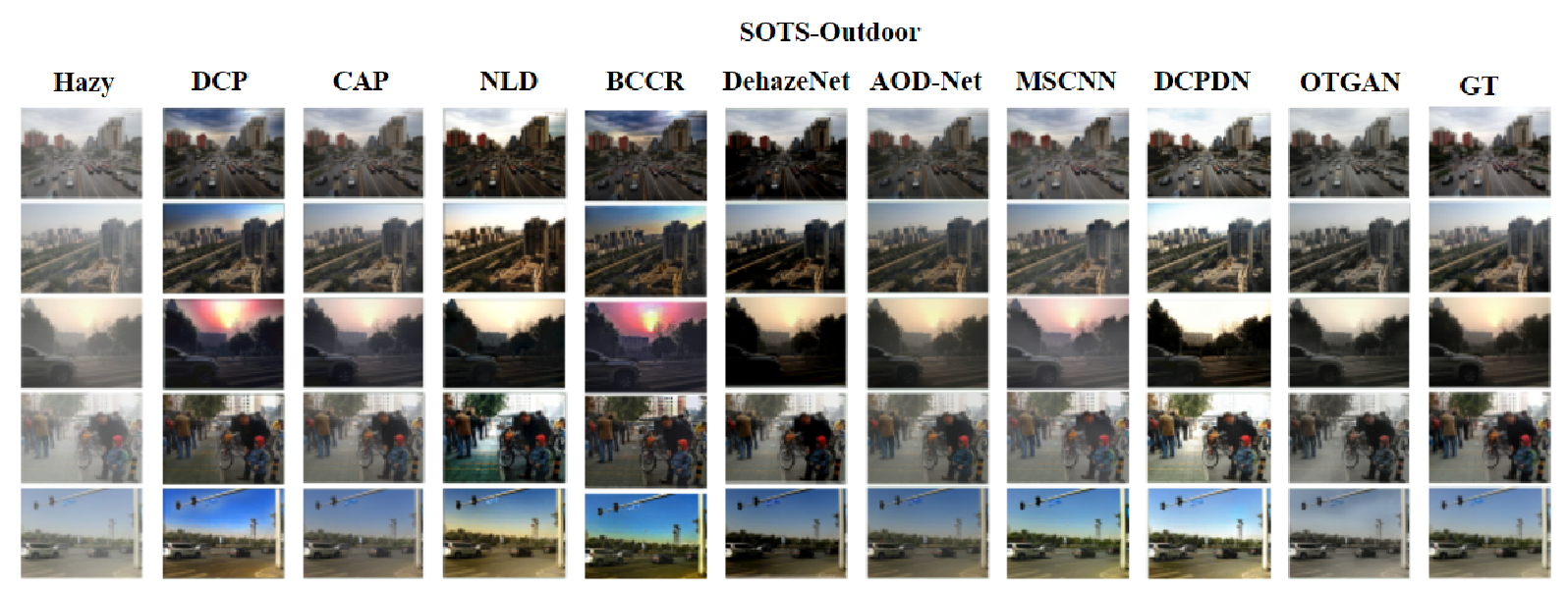}
    \caption{Qualitative comparisons of the dehazed images obtained from the selected approaches as highlighted on top of each column on the SOTS-Outdoor. The second-last column denote the output obtained from the OTGAN, proposed by Kumar et.al. \cite{kumar2022orthogonal}, from which the images is adapted.}
    \label{SOTS_Outdoor_Img}
\end{figure}

\begin{figure}[hbt!]
    \centering
    \includegraphics[scale=0.60]{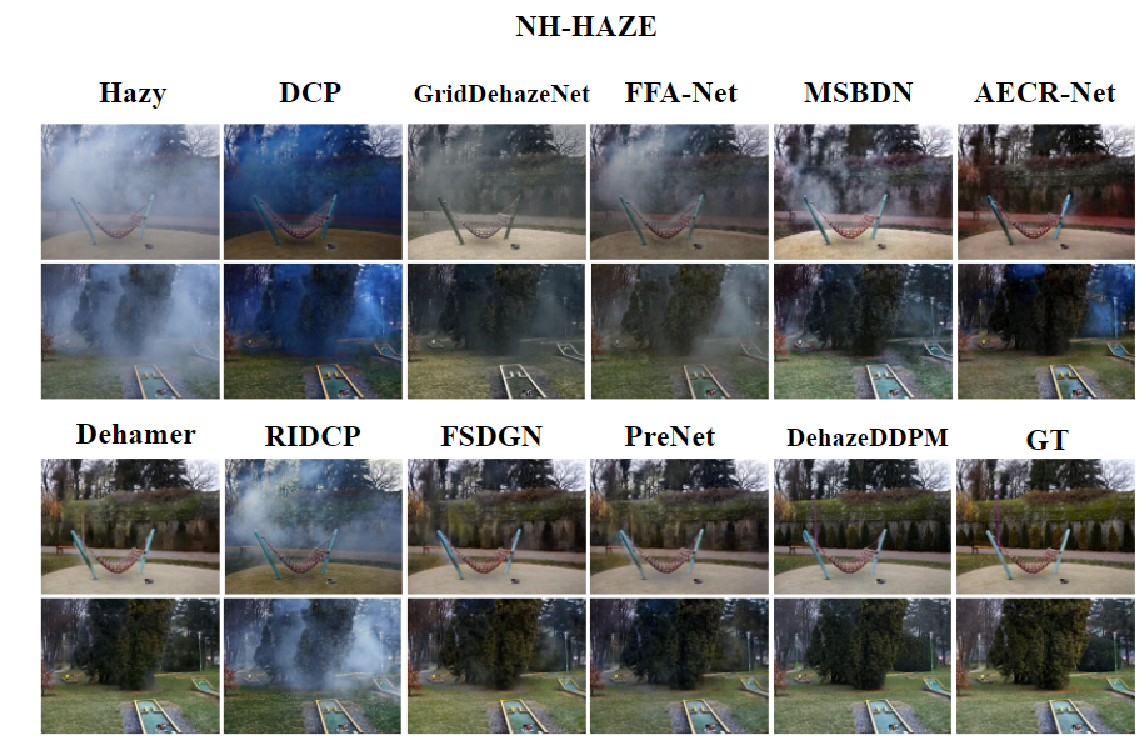}
    \caption{Qualitative comparisons of the dehazed images obtained from the selected approaches as highlighted on top of each column on the NH-HAZE. The RIDCP is from Yu et.al. \cite{yu2022frequency}, the FSDGN is adapted from Wu et.al. \cite{wu2023ridcp}, the MSBDN is from Dong et.al. \cite{dong2020multi}, and the DehazeDDPM is adapted from Yu et.al. \cite{yu2023high}, in which the images are from. PreNet is simply the first stage of the DehazeDDPM, as it comes in two stages.}
    \label{NH_HAZE_Img}
\end{figure}

\begin{figure}[hbt!]
    \centering
    \includegraphics[scale=0.60]{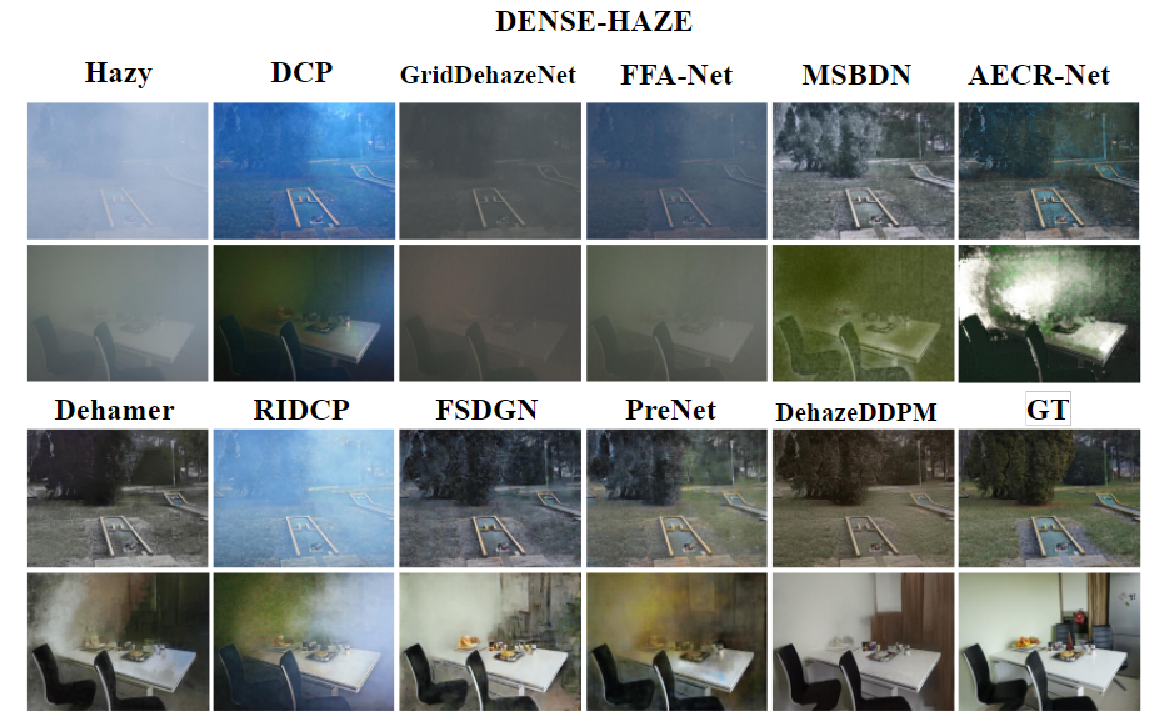}
    \caption{Qualitative comparisons of the dehazed images obtained from the selected approaches as highlighted on top of each column on the DENSE-HAZE. The methods are the same as that presented in the previous figure for the NH-HAZE. The images are adapted from Yu et.al. \cite{yu2023high}.}
    \label{DENSE_HAZE_Img}
\end{figure}

\begin{figure}[hbt!]
    \centering
    \includegraphics[scale=0.60]{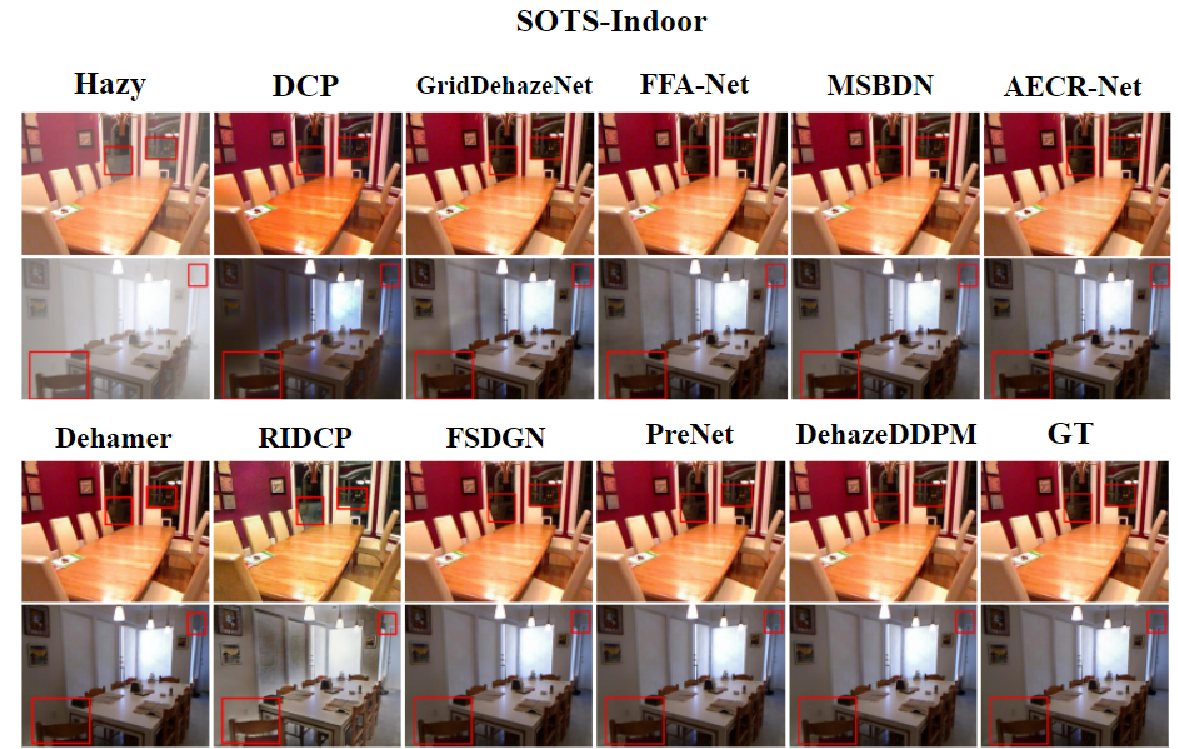}
    \caption{Qualitative comparisons of the dehazed images obtained from the selected approaches as highlighted on top of each column on the SOTS-Indoor. The methods are the same as that presented in the previous figures for the NH-HAZE and DENSE-HAZE. The images are adapted from Yu et.al. \cite{yu2023high}.}
    \label{SOTS_Indoor_Img}
\end{figure}

\begin{figure}[hbt!]
    \centering
    \includegraphics[scale=0.60]{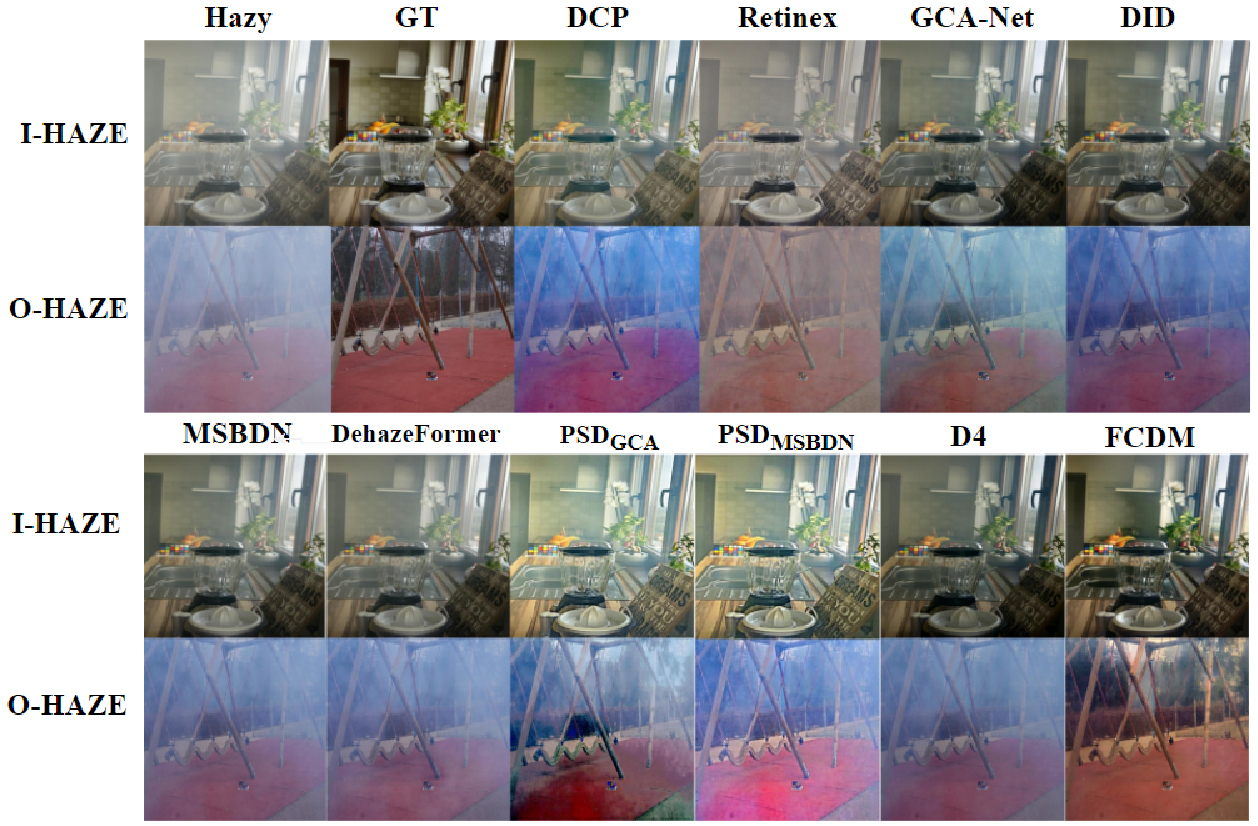}
    \caption{Qualitative comparisons of the dehazed images obtained from the selected approaches as highlighted on top of each column on the I-HAZE and O-HAZE. The images are adapted from Wang et.al. \cite{wang2023frequency}, who proposed the FCDM. Retinex is from Rahman et.al. \cite{rahman1996multi}, DID is from Liu et.al. \cite{liu2021synthetic}, and both $PSD_{GCA}$ and $PSD_{MSBDN}$ are from Chen et.al. \cite{chen2021psd}.}
    \label{I_HAZE_O_HAZE_Img}
\end{figure}

In this appendix, we illustrate some dehazed outputs from the selected approaches for qualitative comparisons. Figure \ref{SOTS_Outdoor_Img}, \ref{NH_HAZE_Img}, \ref{DENSE_HAZE_Img}, \ref{SOTS_Indoor_Img} and \ref{I_HAZE_O_HAZE_Img} illustrates these on the SOTS-Outdoor, NH-HAZE, DENSE-HAZE, SOTS-Indoor, and O-HAZE with I-HAZE, respectively. 

 \bibliographystyle{elsarticle-num} 
 \bibliography{ref}





% \end{thebibliography}
\end{document}